\documentclass[10pt,twocolumn,letterpaper]{article}

\usepackage[pagenumbers]{cvpr} 

\usepackage{indentfirst}
\usepackage{multirow}
\usepackage[normalem]{ulem}
\useunder{\uline}{\ul}{}
\usepackage{graphicx}

\usepackage{bbding} 
\usepackage[inkscapelatex=false]{svg} 
\usepackage[accsupp]{axessibility}
\usepackage{color}

\usepackage{utfsym}
\usepackage[accsupp]{axessibility}  

\definecolor{cvprblue}{rgb}{0.21,0.49,0.74}
\usepackage[pagebackref,breaklinks,colorlinks,allcolors=cvprblue]{hyperref}

\newcommand{\boldparagraph}[1]{\noindent\textbf{#1}\ }

\def\paperID{5656} 
\def\confName{CVPR}
\def\confYear{2025}

\title{MonoDGP: Monocular 3D Object Detection with Decoupled-Query and Geometry-Error Priors}

\author{
    Fanqi Pu$^1$ \qquad
    Yifan Wang$^1$ \qquad
    Jiru Deng$^1$ \qquad
    Wenming Yang$^1$\thanks{Corresponding author.} \\
    $^1$Shenzhen International Graduate School, Tsinghua University \\
    {\tt\small \{pfq23, yf-wang23, djr23\}@mails.tsinghua.edu.cn, yang.wenming@sz.tsinghua.edu.cn}
}

\begin{document}
\maketitle
\begin{abstract}
Perspective projection has been extensively utilized in monocular 3D object detection methods. It introduces geometric priors from 2D bounding boxes and 3D object dimensions to reduce the uncertainty of depth estimation. However, due to errors originating from the object's visual surface, the bounding box height often fails to represent the actual central height, which undermines the effectiveness of geometric depth. Direct prediction for the projected height unavoidably results in a loss of 2D priors, while multi-depth prediction with complex branches does not fully leverage geometric depth. This paper presents a Transformer-based monocular 3D object detection method called MonoDGP, which adopts perspective-invariant geometry errors to modify the projection formula. We also try to systematically discuss and explain the mechanisms and efficacy behind geometry errors, which serve as a simple but effective alternative to multi-depth prediction. Additionally, MonoDGP decouples the depth-guided decoder and constructs a 2D decoder only dependent on visual features, providing 2D priors and initializing object queries without the disturbance of 3D detection. To further optimize and fine-tune input tokens of the transformer decoder, we also introduce a Region Segmentation Head (RSH) that generates enhanced features and segment embeddings. Our monocular method demonstrates state-of-the-art performance on the KITTI benchmark without extra data. Code is available at \url{https://github.com/PuFanqi23/MonoDGP}. 
\end{abstract}    
\vspace{-0.4cm}
\section{Introduction}
\label{sec:introduction}

\begin{figure}[t]
   \centering
   \includegraphics[width=1.0\linewidth]{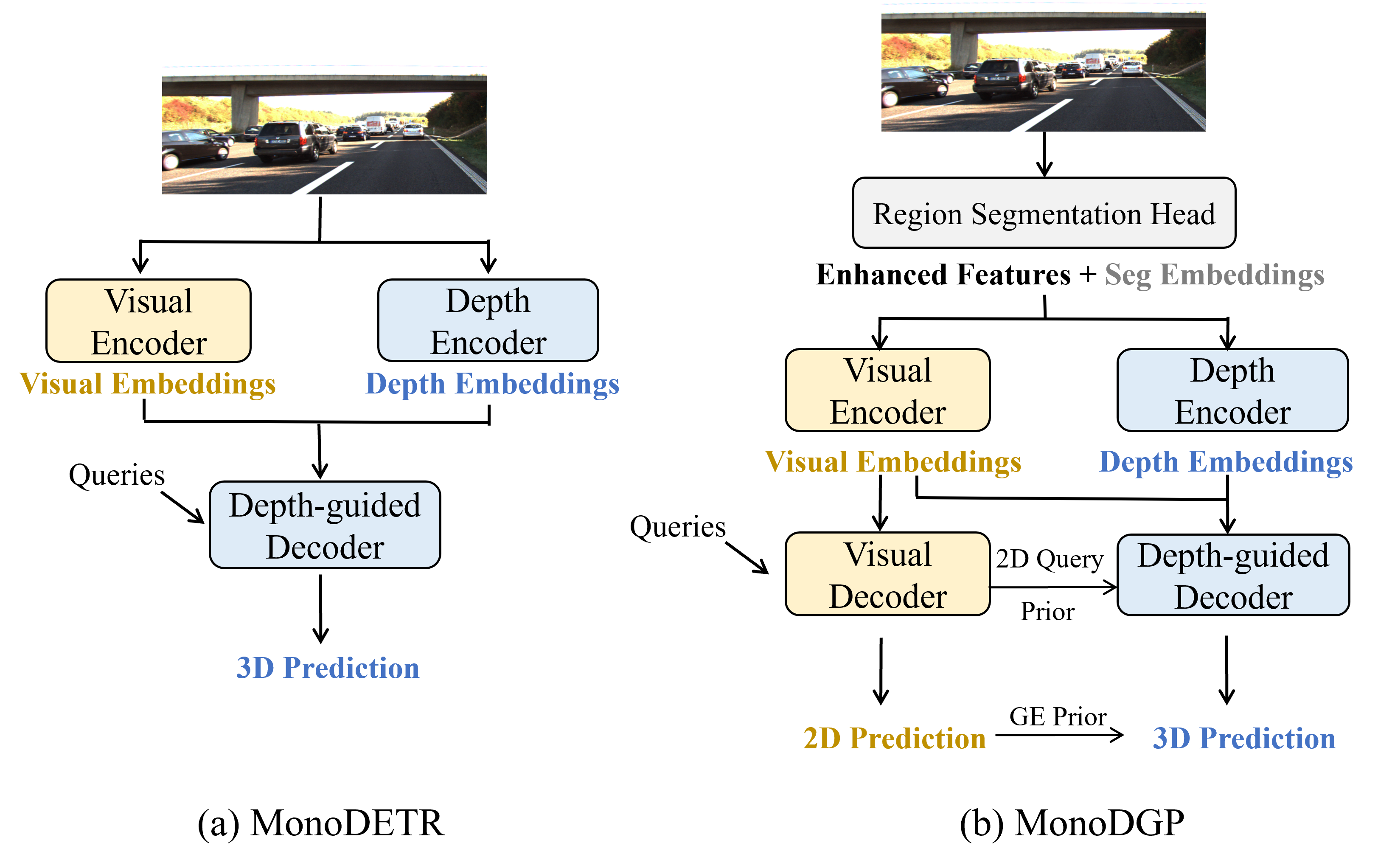}
   \includegraphics[width=1.0\linewidth]{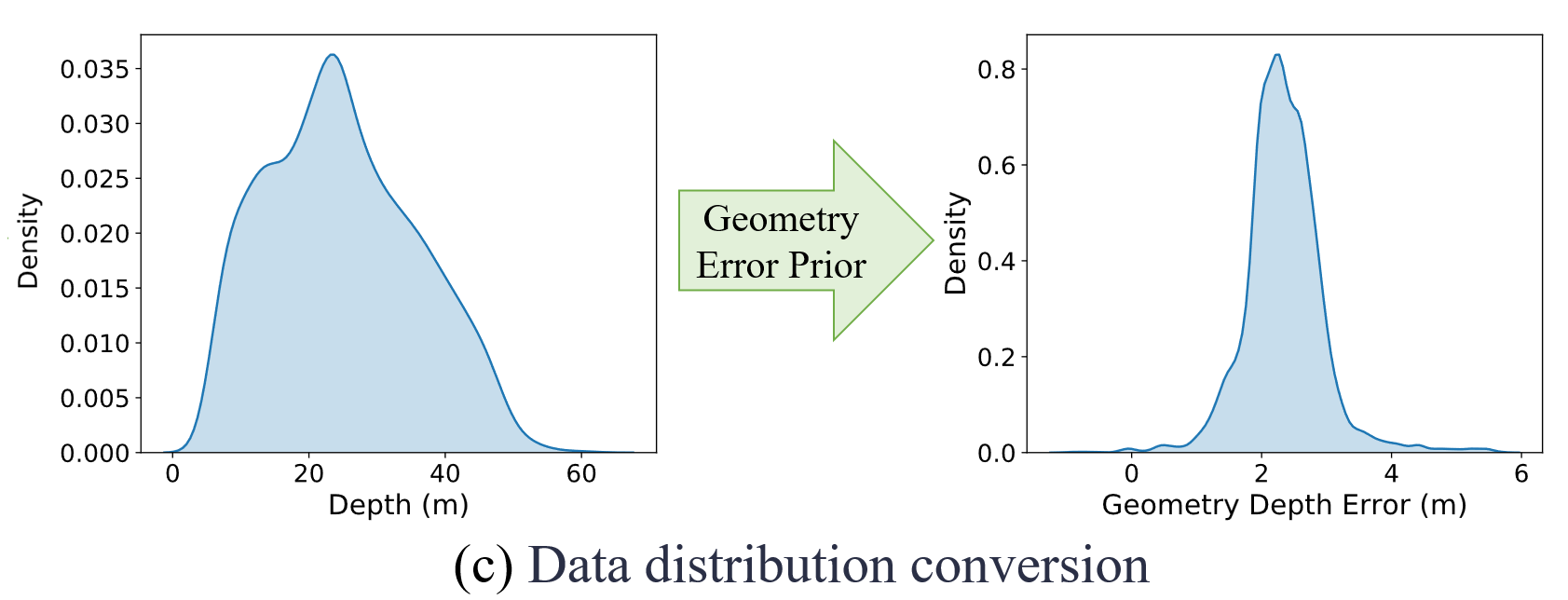}

   \caption{\textbf{Comparison with MonoDETR.} MonoDGP employs an RSH module for enhanced features and segment embeddings, along with an independent visual decoder for 2D query initialization. We further propose a geometry error prior that converts uneven depth distribution into more concentrated error distribution.}
   \label{fig:1}
   \vspace{-15pt}
\end{figure}

3D object detection has always been a critical research topic in the fields of autonomous driving and robotic perception. Compared with LiDAR- \cite{PointPillar,PV-RCNN,PV-RCNN++,BtcDet} or stereo-based \cite{rts3d,stereo_centernet,side,Stereo_R-CNN, dsgn++} methods, monocular 3D object detection exhibits lower spatial positional accuracy without depth information. Owing to their cost-effectiveness and ease of configuration~\cite{monodde,survey}, monocular methods have remained a primary focus of research in recent years.

Without extra data, the direct prediction of 3D objects from single 2D images is an extremely challenging problem. During the formation of 2D images, crucial depth-related information is compressed, resulting in an inherently ill-posed issue for direct depth estimation~\cite{monoground,monoedge,DID-M3D,monoflex,monodde}. This limitation significantly undermines the accuracy of 3D object detection. However, under certain conditions, the uncertainty of depth estimation can be reduced with substantial prior knowledge. Particularly in the realm of autonomous driving, where the major objects of interest, such as cars, are rigid entities with known geometric properties. 

Previous methods like Deep3DBox~\cite{deep3dbox} and Shift R-CNN~\cite{shift-rcnn} enforce geometric constraints through tight alignment between 3D box projections and 2D detections. Unlike strict constraints mentioned above, geometric depth is uniquely determined by the ratio of an object's 3D height to its 2D projected height. This physically interpretable explicit mapping eliminates cascaded uncertainty, ensuring superior computational efficiency and noise robustness. However, since 2D images are inaccurate in capturing depth errors of the object visual surface, the 2D bounding box height often exceeds the projected height, which makes geometric depth deviate from the actual depth. To address single-depth prediction limitations, some approaches\cite{monodetr, monodde, monocd} integrate multi-depth predictions through weighted averaging, including direct and geometric depths. Nevertheless, MonoCD~\cite{monocd} indicates that if these predicted depths cluster on the same side of the ground truth depth, even weighted averaging fails to correct the depth error. DID-M3D~\cite{DID-M3D} decomposes depth into visual and attribute parts, using perspective transformation for data augmentation. Similarly, the geometric depth can also be regarded as the visual depth, whose value depends on the object's surface appearance and position \cite{neuralseedepth}. While the depth error is determined by the object's inherent attributes and remains invariant with changes in perspective. With more concentrated distribution, error prediction can not only alleviate the long-tail effect of raw depth, but also avoid the problem where multiple depths fail to complement each other after weighting. 

After introducing the geometric prior, we observed that the accuracy of the bounding boxes is often poor during the initial training stage. This may lead the model to associate depth with incorrect bounding boxes, thereby complicating the convergence process. To address this issue, we decouple the depth-guided decoder proposed by MonoDETR~\cite{monodetr} and construct a 2D visual decoder for independent 2D detection and query initialization. Additionally, we propose a Region Segmentation Head (RSH) to strengthen the foreground features and provide segment embeddings for semantically similar input tokens. By emulating BERT~\cite{bert}'s approach to handling distinct components of textual inputs, we treat the target region as one “sentence” and the background region as another “sentence”.

In summary, we propose a transformer-based method called MonoDGP. Our contributions are listed as follows:

\begin{itemize} 
   \item We adopt perspective-invariant error prediction to replace multi-depth prediction, alleviating the long-tail effect of depth distribution and reducing learning complexities.\vspace{0.1cm}
   
   \item We construct a decoupled visual decoder to initialize queries and reference points of the depth-guided decoder, boosting the model convergence and training stability.\vspace{0.1cm}
   
   \item We present a Region Segmentation Head (RSH) to enhance object features and suppress background noise. Additionally, segment embeddings are incorporated during depth encoding to refine contextual representations.\vspace{0.1cm}
   
   \item Evaluated on the KITTI 3D object detection benchmark, without any extra data, MonoDGP achieves the state-of-the-art (SOTA) performance among monocular detectors.
   
\end{itemize}

\begin{figure*}
   \centering
   
   \includegraphics[width=0.80\linewidth]{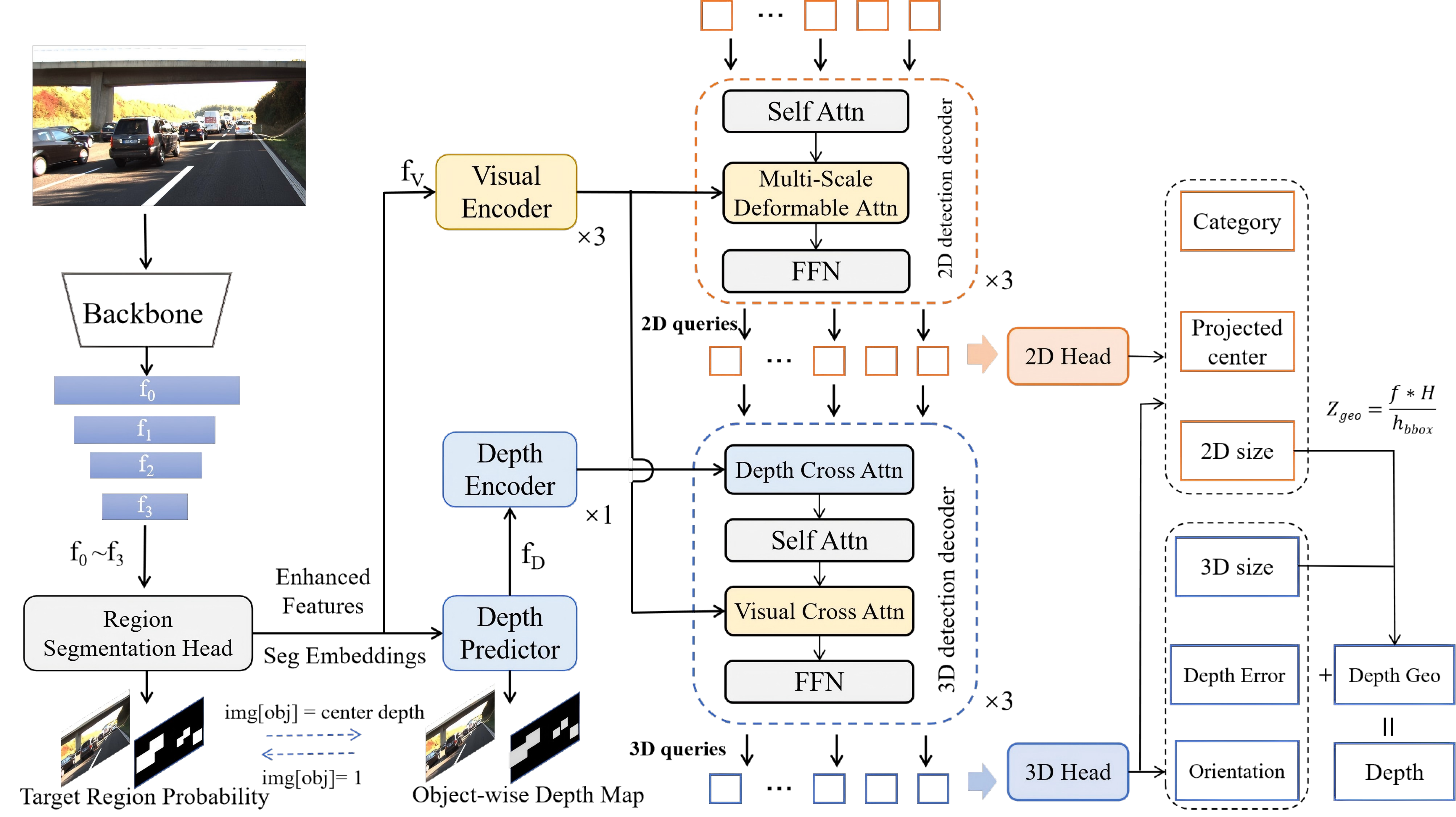}
   \caption{\textbf{Overall architecture of our MonoDGP.} The network comprises three components: feature extraction and enhancement, transformer encoder-decoder and object detection heads. The parallel visual and depth branches within the transformer are represented by {\color[HTML]{FFD700} yellow} and {\color[HTML]{3399ff} blue}, respectively. The predicted depth is separated into geometric depth $Z_{geo}$, obtained from the perspective projection formula, and depth error $Z_{err}$ to correct the formula.}
   \label{fig:2}
\end{figure*}

\section{Related work}
\label{sec:related}

\boldparagraph{Extra training data.}
 The main challenge of monocular methods is insufficient of depth cues, which renders 3D object detection an ill-posed problem. Consequently, some approaches \cite{ddmp-3d, caddn, DCD, monodtr, DID-M3D, monopgc} incorporate additional data, such as point clouds or depth maps, enabling the model to implicitly learn depth features during training. MonoPGC~\cite{monopgc} introduces pixel-level depth map prediction as an auxiliary task, encoding 3D coordinates into depth-perception features to perceive the spatial positions of pixels. CaDDN~\cite{caddn} estimates the classified depth distribution for each pixel, generates high-quality bird's-eye view feature representations, and achieves accurate 3D object predictions. OccupancyM3D~\cite{occupancym3d} creates occupancy labels from point clouds, and monitors occupancy status in frustum space and 3D space. Given the reliance on expensive sensors and complex collection processes for supplementary information, MonoDGP exclusively utilizes monocular images during the training and inference stages, thereby enhancing the model's practicality and deployment feasibility.

\boldparagraph{Transformer-based methods.}
 Many previous works \cite{monoflex, monorcnn, gupnet, DID-M3D, monocon, monodde} have employed convolutional neural networks (CNNs) to extract local, fine-grained information at various spatial levels. Although CNN-based methods demonstrate superior speed and efficiency, they require multi-layer convolution to progressively aggregate features for capturing long-distance dependencies. Additionally, non-maximum suppression (NMS) technology is employed to filter out redundant detection boxes. Benefiting from the global perception and dynamic adjustment abilities of the attention mechanism~\cite{attention}, an increasing number of studies \cite{monodtr, monodetr, monoatt} have shifted their focus toward Transformer-based monocular detectors. For instance, MonoDTR~\cite{monodtr} proposes to conduct depth position encoding and integrates global depth information into the Transformer to guide the detection. MonoDETR~\cite{monodetr} adheres to the DETR~\cite{detr} architecture, utilizing object-level depth labels to predict the foreground depth map, constructing a depth-guided decoder, and incorporating object queries for global feature aggregation. MonoDGP also adopts the DETR framework, initializing queries with a decoupled 2D visual decoder and introducing segment embeddings to enhance the contextual relationships among pixels.

\boldparagraph{Multi-depth estimation modes.}
 Early monocular methods \cite{chen2016monocular, simonelli2019disentangling, Kinematic3D, liu2020reinforced, smoke} directly estimate object depth through end-to-end network training. Many researchers \cite{bao2020object, barabanau2019monocular, cai2020monocular, zhou2021monocular, monopsr, monorcnn, wang2022probabilistic} extend the depth estimation branches by introducing perspective projection and mathematical priors. For example, GUPNet~\cite{gupnet} proposes a geometry uncertainty projection module and employs a Laplacian distribution to represent geometric depth with correlation-learned bias. To mitigate uncertainties in single depth estimation methods, MonoDDE~\cite{monodde} introduces key points into 20 different depth branches, emphasizing the significance of diversity in depth prediction. MonoDETR~\cite{monodetr} merges predicted results from direct depth, object-level depth map, and geometric depth. MonoCD~\cite{monocd} further discusses the complementarity among multiple depths and combines them to cancel out positive or negative errors. However, the geometric depth and errors (such as depth/height error) in the projection formula have not been fully explored. Based on a comprehensive analysis of geometric priors, MonoDGP decouples raw depth to obtain perspective-invariant depth error, providing a superior alternative to multi-depth approaches.
\section{Method}
\label{sec:method}

\begin{figure*}
   \centering
   
   \includegraphics[width=0.75\linewidth, height=7.0cm]{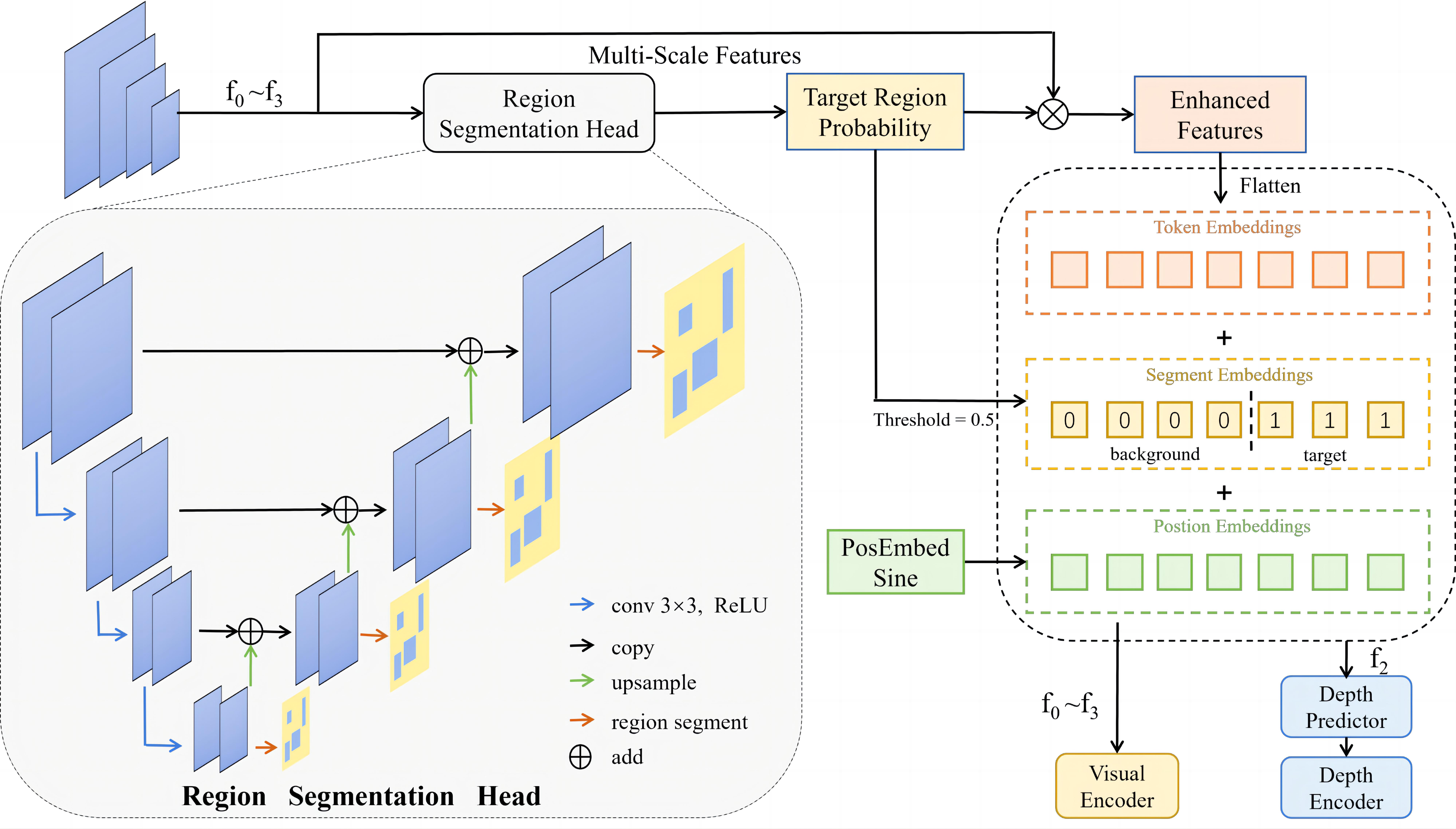}
   \caption{\textbf{The structure of region segmentation module.} The multi-scale feature maps are progressively upsampled and added, outputting target region probabilities by segment heads. Subsequently, enhanced features are achieved through element-wise multiplication with original features. Finally, segment embeddings are acquired under threshold constraints.}
   \label{fig:3}
\end{figure*}

\subsection{Overview}

The overall framework of MonoDGP is shown in \cref{fig:2}. Firstly, a single-view image is processed by ResNet50 to extract multi-scale feature maps \({f_i} \in {\mathbb{R}^{\frac{H}{{{2^{i + 3}}}} \times \frac{W}{{{2^{i + 3}}}} \times C}},i = 0,1,2,3\). The RSH module then predicts target region probabilities and enhances visual features $f_{V}$. A lightweight depth predictor subsequently estimates the depth map and generates depth features $f_{D}$. Both enhanced visual features $f_{V}$ and depth features $f_{D}$ are fed into two parallel transformer encoders. The transformer decoder consists of a 2D visual decoder and a 3D depth-guided decoder. In the 2D decoder, object queries adaptively achieve semantic scene understanding from visual embeddings, while the 3D decoder extracts geometric and spatial relationships from both visual and depth embeddings. Ultimately, we utilize decoupled queries as inputs for 2D/3D detection heads separately. It is worth noting that after calculating geometric depth based on the projection formula, a geometry error correction mechanism is applied to improve the accuracy of depth estimation.

\subsection{Region Segmentation}

The Region Segmentation Head (RSH) is specifically designed to enhance the fine-grained foreground feature extraction. As shown in \cref{fig:3}, the RSH follows the structure of U-Net~\cite{unet} and predicts the target probability for each pixel. It is essential for the depth predictor to concentrate more on the perspective and geometric cues within the foreground area. Therefore, the RSH provides enhanced target features and segment embeddings for depth encoding. Technical details are presented below:

\boldparagraph{Region Segmentation and Enhancement.} 
We initially utilize a ResNet backbone to downsample and extract multi-scale visual features \( f_0 \) to \( f_3 \).  Subsequently, We restore \( f_3 \) to high-resolution feature maps through upsampling operations and skip connections. This integration of shallow and deep features improves multi-scale representation. Specifically, we combine the upsampled \( f_3 \) with \( f_2 \) after a \( 1 \times 1 \) convolution and GroupNorm layer. This process is repeated to obtain fused feature maps at various scales. We then employ an SEBlock~\cite{seblock} for attention weighting on each scale's feature maps. The SEBlock implements a channel attention mechanism through two \( 1 \times 1 \) convolution layers: the first convolution reduces the number of feature channels to \( 1/16 \),  while the second convolution restores the original number of channels. Finally, we achieve several semantic segmentation maps with the sigmoid function. By multiplying the predicted target probability with original features, RSH effectively enhances foreground features and suppresses background noise.

\begin{figure*}
   \centering
   \includegraphics[width=0.95\linewidth]{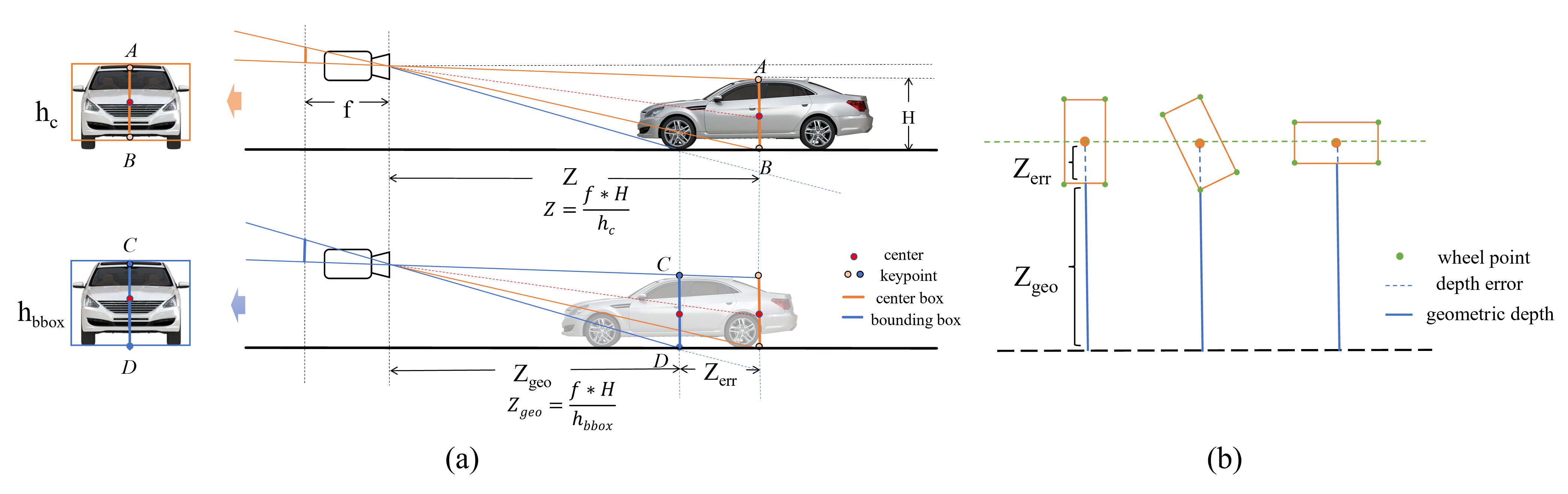}
   \caption{(a) \textbf{Schematic illustration of how geometry error occurs.} Due to the height of the bounding box \( h_{\text{bbox}} \) is larger than the projected central height \( h_c \), there will be an error between geometric depth and ground truth depth.
   (b) \textbf{Vehicles at different angles in a bird's-eye view.} The geometric depth \( Z_{\text{geo}} \) is the distance between the camera plane and the parallel plane on which the closest wheel sits, while depth error \( Z_{\text{err}} \) only depends on the object's dimensions and orientation. The camera height also has a tiny effect on the \( Z_{\text{err}} \) , rigorous proof is given in the appendix.
   }
   \label{fig:4}
\end{figure*}

\boldparagraph{Region loss.} 
We construct target region masks and set the areas within bounding boxes to 1. Guided by these region masks, we calculate the Dice loss for the target region probabilities at each scale and weight the sum of these losses. The Dice loss, which is based on set similarity, addresses class imbalance and improves segmentation precision. Its definition is as follows:
\begin{equation}
L_{region}^i = 1 - \frac{{2 \cdot \sum\limits_{j = 1}^{{N_i}} {{p_j} \cdot {g_j}} }}{{\sum\limits_{j = 1}^{{N_i}} {{p_j}}  + \sum\limits_{j = 1}^{{N_i}} {{g_j}} }}
\end{equation}
Where \( p_j \) denotes the predicted probability  of the \( j \)-th pixel, \( g_j \) represents the \( j \)-th pixel in the ground truth label, and \( N_i \) indicates the total number of pixels in the \( i \)-th feature map \({f_i} ,i=0,1,2,3\). By minimizing the Dice loss, we can directly optimize the overlapping areas between the predicted probabilities and the region masks, enhancing both segmentation accuracy and consistency.

\boldparagraph{Segment embeddings.} 
The RSH module generates segment embeddings from target region probabilities that exceed a threshold of 0.5. We subsequently add these segment embeddings to the token and position embeddings. This approach is inspired by the segment embeddings mechanism utilized in the BERT model~\cite{bert}. Segment embeddings help to better distinguish the semantic relationships between foreground and background image tokens, similar to how different “sentences” are represented in BERT.

\subsection{Decoupled Query}
\label{sec:detector}
Geometric priors optimize depth estimation, but also introduce depth uncertainty to 2D detection. To mitigate this side effect, we decouple object queries for independent 2D detection. After obtaining visual embeddings \( f_V^e \) and depth embeddings \( f_D^e \) from two parallel encoders, MonoDGP decomposes the depth-guided decoder into a 2D decoder and a 3D decoder. The 2D visual decoder initialize 2D queries and reference points to acquire spatial position and category information.
These 2D queries are then fed into the 3D decoder to interact with visual and depth features. This allows each decoder to focus on its own specific task and ensures stable convergence during training.

\boldparagraph{2D detection decoder.} 
We take a set of learnable object queries \(q \in {\mathbb{R}^{N \times C}}\) as input, which sequentially passes through an inter-query self-attention layer, a multi-scale deformable cross-attention layer~\cite{deform-detr}, and a feedforward network(FFN), where \(N\) denotes the pre-defined maximum number of objects in a single image. The 2D queries \({q_{2d}} \) interact with visual embeddings \( f_V^e\) to extract 2D image features, including color, edges, and texture. The queries and reference points provide an initial perception of spatial positioning. The 3D decoder can utilize these 2D query priors to more rapidly focus on the target area in complex scenes.

\boldparagraph{3D detection decoder.} 
The initialized 2D object queries \({q_{2d}} \in {R^{N \times C}}\) are input to the 3D depth-guided decoder. Each decoder block contains a depth cross-attention layer, an inter-query self-attention layer, a visual cross-attention layer and an FFN. Specifically, object queries first extract perspective cues from depth embeddings \( f_D^e\). Subsequently, the depth-aware queries are sent to the inter-query self-attention layer to promote information exchange. After the visual cross-attention layer and FFN, we generate 3D queries \({q_{3d}}\) that combine semantic and geometric features, obtaining an accurate scene-level spatial understanding.

\subsection{Geometry Error}

Direct depth estimation is often ill-posed in monocular methods. We apply the perspective projection formula \( Z = f \times H_{3D} / h_{2D} \) to introduce geometric priors and mitigate depth uncertainty. When utilizing the projection formula, it is essential to make sure that the object's dimension height (such as \( AB \) and \( CD \) in \cref{fig:4}(a)) is parallel to the projection plane, namely the X-Y plane in the camera coordinate system, and remains perpendicular to the X-Z plane. To guarantee the validity of the formula, we establish the following assumptions:
\begin{itemize} 
\item The roll and pitch angles of the monocular camera are set to 0 degrees by default.
\item The influence of ground slope can be overlooked when calculating geometric depth.
\end{itemize} 

Under the given assumptions, the projected central height \( h_c \) is solely dependent on the 3D height and depth, while the height of the 2D bounding box \( h_{\text{bbox}} \) can be influenced by additional factors, including its shape and surface characteristics. As illustrated in \cref{fig:4} (a), the visual disparity typically causes a depth error \( Z_{\text{err}} \) between the actual depth \( Z \) and the geometric depth \( Z_{\text{geo}} \), which can be formulated as:
\begin{equation}
Z = \frac{{f \cdot {Y_{AB}}}}{{{v_{AB}}}} = \frac{{f \cdot H}}{{{h_c}}},{Z_{geo}} = \frac{{f \cdot {Y_{CD}}}}{{{v_{CD}}}} = \frac{{f \cdot H}}{{{h_{bbox}}}}
\label{eq:2}
\end{equation}
\begin{equation}
Z = {Z_{geo}} + {Z_{err}}
\label{eq:3}
\end{equation}
where \( f\) denotes the focal length of the camera.

The \( Z_{\text{geo}} \) reflects the visible surface depth, while \( Z_{\text{err}} \) is only related to the inherent attributes of the object. In comparison to the raw depth \( Z \), \( Z_{\text{err}} \) is perspective-invariant with less spatial uncertainty. We replace the direct depth prediction with the geometry error prediction, and our depth loss is calculated as follows:
\begin{equation}
{L_{depth}} = \frac{{\sqrt 2 }}{{{\sigma _d}}}{\left\| {{Z_{geo}} + {Z_{err}} - {Z_{gt}}} \right\|_1} + \log ({\sigma _d})
\label{eq:4}
\end{equation}

The perspective projection formula contains two variables: the height of the 2D bounding box and the height of the 3D size. To correct the formula, the depth error is added to the geometric depth for the final prediction, while errors from both the 2D and 3D heights can also compensate for the geometric depth. Their predictions are formulated as:
\begin{equation}
Z = \frac{{f \cdot (H + {H_{err}})}}{{{h_{bbox}}}} = {Z_{geo}} + \frac{f}{{{h_{bbox}}}} \cdot {H_{err}}
\label{eq:5}
\end{equation}
\begin{equation}
Z = \frac{{f \cdot H}}{{{h_c}}} = \frac{{f \cdot H}}{{{h_{bbox}} - {h_{err}}}}
\label{eq:6}
\end{equation}

These errors can significantly improve the performance of geometric depth.

\begin{figure}
   \centering
   \includegraphics[width=1.0\linewidth]{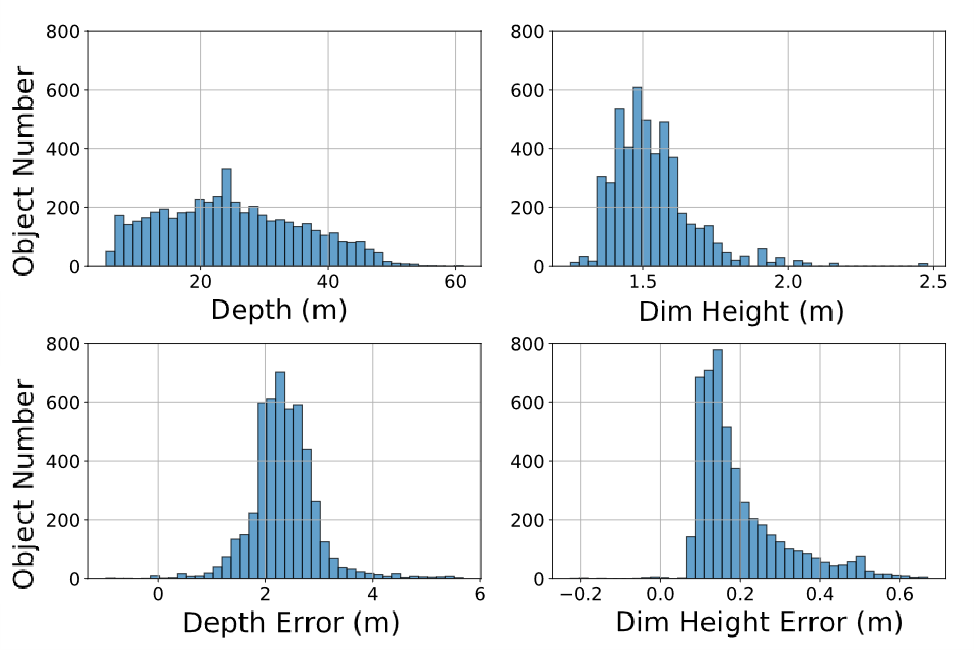}
   \caption{\textbf{Comparison of 3D data distributions.} With the total number of objects remaining constant, each distribution is visualized using a histogram.}
   \label{fig:5}
\end{figure}

\subsection{Why Geometry Error Works?}

The geometry error eliminates perspective correlation of raw depth distribution, thus reducing the learning difficulty of the network. Specifically, the reasons why geometry error works can be analyzed from three aspects: \vspace{0.15cm}
\begin{itemize} 
\item {2D and 3D attributes}: why we utilize \(h_{bbox} \) instead of \(h_c \) in the projection formula?\vspace{0.15cm}
\item {Different 3D attributes}: why raw depth preforms worse than other 3D attributes?\vspace{0.15cm}
\item {Different geometry errors}: why we select depth error?\vspace{0.2cm}
\end{itemize} 

\boldparagraph{2D and 3D attributes.}
2D predictions rely on monocular visual features like color, edges, textures, whereas 3D predictions require extra factors including depth positioning and geometric perspective, which are difficult to obtain from single-view images. Benefiting from the network’s powerful capability of marginal feature extraction, \(h_{bbox} \) can provide more accurate 2D priors. However, we can hardly predict \(h_c \) from direct marginal features. The geometric depth calculated by \(h_{bbox}\) should be regarded as a reliable surface depth, rather than an inaccurate central depth.

\begin{figure}
   \centering
   \includegraphics[width=1.0\linewidth]{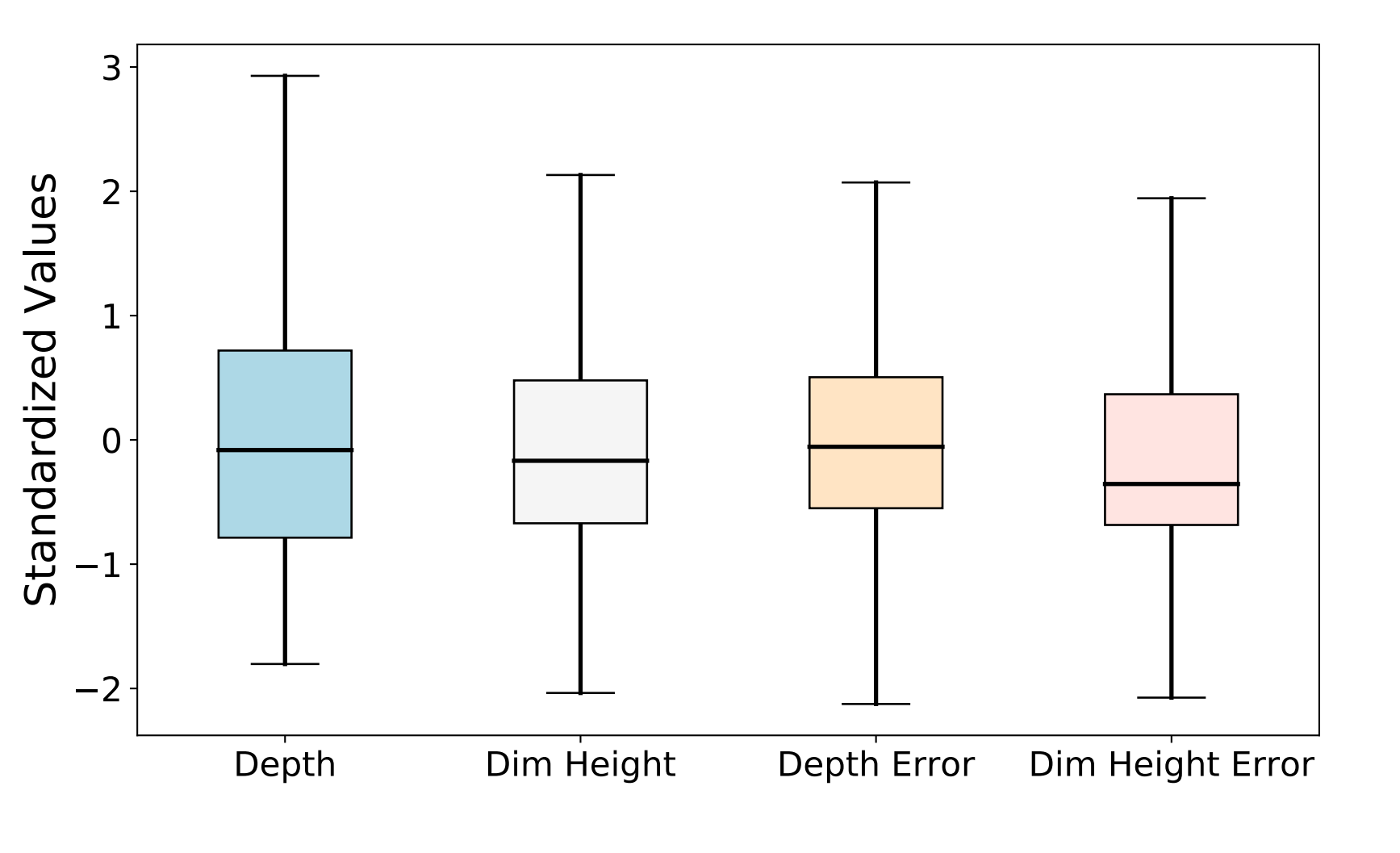}
   \caption{\textbf{The boxplot of depth-related attributes.} To conduct unified comparisons, the data distributions of different attributes are standardized by \(X' = (X - \mu )/\sigma \) in the boxplot.}
   \label{fig:6}
\end{figure}

\begin{table*}[!ht]
\centering

\resizebox{\textwidth}{30mm}{
\begin{tabular}{l|c|c|ccc|clcc|clcc|clcc}
\hline
\multicolumn{1}{c|}{} &  &  & \multicolumn{3}{c|}{Test, $AP_{BEV|R40}$} & \multicolumn{4}{c|}{Test, $AP_{3D|R40}$} & \multicolumn{4}{c|}{Val, $AP_{BEV|R40}$} & \multicolumn{4}{c}{Val, $AP_{3D|R40}$} \\ \cline{4-18} 
\multicolumn{1}{l|}{\multirow{-2}{*}{Methods}} & \multirow{-2}{*}{Extra data} & \multirow{-2}{*}{Reference} & Easy & Mod. & Hard & \multicolumn{2}{c}{Easy} & Mod. & Hard & \multicolumn{2}{c}{Easy} & Mod. & Hard & \multicolumn{2}{c}{Easy} & Mod. & Hard \\ \hline
CaDDN~\cite{caddn} &  & CVPR 2021 & 27.94 & 18.91 & 17.19 & \multicolumn{2}{c}{19.17} & 13.41 & 11.46 & \multicolumn{2}{c}{-} & - & - & \multicolumn{2}{c}{23.57} & 16.31 & 13.84 \\
MonoDTR~\cite{monodtr} &  & CVPR 2022 & 28.59 & 20.38 & 17.14 & \multicolumn{2}{c}{21.99} & 15.39 & 12.73 & \multicolumn{2}{c}{33.33} & 25.35 & 21.68 & \multicolumn{2}{c}{24.52} & 18.57 & 15.51 \\
DID-M3D~\cite{DID-M3D} &  & ECCV 2022 & 32.95 & 22.76 & 19.83 & \multicolumn{2}{c}{24.40} & 16.29 & 13.75 & \multicolumn{2}{c}{31.10} & 22.76 & 19.50 & \multicolumn{2}{c}{22.98} & 16.12 & 14.03 \\
OccupancyM3D~\cite{occupancym3d} & \multirow{-4}{*}{LiDAR} & CVPR 2024 & \textbf{35.38} & \underline{24.18} & \underline{21.37} & \multicolumn{2}{c}{\underline{25.55}} & 17.02 & 14.79 & \multicolumn{2}{c}{35.72} & 26.60 & \underline{23.68} & \multicolumn{2}{c}{26.87} & 19.96 & \underline{17.15} \\ \hline
MonoPGC~\cite{monopgc} &  & ICRA 2023 & 32.50 & 23.14 & 20.30 & \multicolumn{2}{c}{24.68} & \underline{17.17} & 14.14 & \multicolumn{2}{c}{34.06} & 24.26 & 20.78 & \multicolumn{2}{c}{25.67} & 18.63 & 15.65 \\
OPA-3D~\cite{opa-3d} & \multirow{-2}{*}{Depth} & RAL 2023 & 33.54 & 22.53 & 19.22 & \multicolumn{2}{c}{24.60} & 17.05 & 14.25 & \multicolumn{2}{c}{33.80} & 25.51 & 22.13 & \multicolumn{2}{c}{24.97} & 19.40 & 16.59 \\ \hline
GUPNet~\cite{gupnet} &  & ICCV 2021 & - & - & - & \multicolumn{2}{c}{20.11} & 14.20 & 11.77 & \multicolumn{2}{c}{31.07} & 22.94 & 19.75 & \multicolumn{2}{c}{22.76} & 16.46 & 13.72 \\
MonoCon~\cite{monocon} &  & AAAI 2022 & 31.12 & 22.10 & 19.00 & \multicolumn{2}{c}{22.50} & 16.46 & 13.95 & \multicolumn{2}{c}{-} & - & - & \multicolumn{2}{c}{26.33} & 19.01 & 15.98 \\
DEVIANT~\cite{deviant} &  & ECCV 2022 & 29.65 & 20.44 & 17.43 & \multicolumn{2}{c}{21.88} & 14.46 & 11.89 & \multicolumn{2}{c}{32.60} & 23.04 & 19.99 & \multicolumn{2}{c}{24.63} & 16.54 & 14.52 \\
MonoDDE~\cite{monodde} &  & CVPR 2022 & 33.58 & 23.46 & 20.37 & \multicolumn{2}{c}{24.93} & 17.14 & \underline{15.10} & \multicolumn{2}{c}{35.51} & 26.48 & 23.07 & \multicolumn{2}{c}{26.66} & 19.75 & 16.72 \\
MonoUNI~\cite{monouni} &  & NeurlPS 2023 & - & - & - & \multicolumn{2}{c}{24.75} & 16.73 & 13.49 & \multicolumn{2}{c}{-} & - & - & \multicolumn{2}{c}{24.51} & 17.18 & 14.01 \\
MonoDETR~\cite{monodetr} &  & ICCV 2023 & 33.60 & 22.11 & 18.60 & \multicolumn{2}{c}{25.00} & 16.47 & 13.58 & \multicolumn{2}{c}{\underline{37.86}} & \underline{26.95} & 22.80 & \multicolumn{2}{c}{\underline{28.84}} & \underline{20.61} & 16.38 \\
MonoCD~\cite{monocd} &  & CVPR 2024 & 33.41 & 22.81 & 19.57 & \multicolumn{2}{c}{25.53} & 16.59 & 14.53 & \multicolumn{2}{c}{34.60} & 24.96 & 21.51 & \multicolumn{2}{c}{26.45} & 19.37 & 16.38 \\
FD3D~\cite{fd3d} & \multirow{-8}{*}{None} & AAAI 2024 & 34.20 & 23.72 & 20.76 & \multicolumn{2}{c}{25.38} & 17.12 & 14.50 & \multicolumn{2}{c}{36.98} & 26.77 & 23.16 & \multicolumn{2}{c}{28.22} & 20.23 & 17.04 \\ \hline
Ours & None & - & \underline{35.24} & \textbf{25.23} & \textbf{22.02} & \multicolumn{2}{c}{\textbf{26.35}} & \textbf{18.72} & \textbf{15.97} & \multicolumn{2}{c}{\textbf{39.40}} & \textbf{28.20} & \textbf{24.42} & \multicolumn{2}{c}{\textbf{30.76}} & \textbf{22.34} & \textbf{19.02} \\
Improvement & - & - & {\color[HTML]{FE0000} -0.14} & {\color[HTML]{3166FF} +1.05} & {\color[HTML]{3166FF} +0.65} & \multicolumn{2}{c}{{\color[HTML]{3166FF} +0.80}} & {\color[HTML]{3166FF} +1.55} & {\color[HTML]{3166FF} +0.87} & \multicolumn{2}{c}{{\color[HTML]{3166FF} +1.54}} & {\color[HTML]{3166FF} +1.25} & {\color[HTML]{3166FF} +1.33} & \multicolumn{2}{c}{{\color[HTML]{3166FF} +1.92}} & {\color[HTML]{3166FF} +1.73} & {\color[HTML]{3166FF} +1.87} \\ \hline
\end{tabular}}
\caption{Comparisons with state-of-the-art monocular methods on the KITTI test and val sets for the car category. We \textbf{bold} the best results and \underline{underline} the second-best results. The {\color[HTML]{3166FF} blue} refers to the gain and the {\color[HTML]{FE0000} red} is the decrease of our method.}
\label{tab:1}
\end{table*}

\boldparagraph{Different 3D attributes.}
Without direct depth supervision, data distribution dictates network learning complexity. The \cref{fig:5} and \ref{fig:6} show the data distributions and boxplot of depth-related 3D attributes. Based on these figures, we can further evaluate and compare the learning difficulty across different data distributions. The central depth exhibits a long-tailed distribution with wide variance, while the 3D size distribution tends to be more concentrated. For instance, typical car height clusters near 1.5 meters, simplifying pattern recognition. The boxplot indicates that the depth distribution has a broader interquartile range (IQR) and longer whiskers, suggesting greater complexity and noise. Meanwhile, the IQR for dimension height, depth error, and height error is narrower, which illustrates that these distributions are relatively compact with plain patterns.

\boldparagraph{Different geometry errors.}
(1) Depth error \(Z_{err} \) and dimension height error \(H_{err} \) :  Both \(Z_{err} \) and \(H_{err} \) have a narrow IQR and short whiskers, indicating that they share similar statistical properties. The primary distinction is that \(Z_{err} \) distribution is symmetrical with a median close to zero, while \(H_{err} \) distribution is skewed with a negative median. This skewness may increase the complexity of the training process. According to \cref{eq:3} and \cref{eq:5}, we can express \(H_{err}\) in terms of \(Z_{err}\) and \(h_{bbox}\): 
\begin{equation}
{H_{err}} = \frac{1}{f} \cdot (Z - {Z_{geo}}) \cdot {h_{bbox}} = \frac{{{Z_{err}} \cdot {h_{bbox}}}}{f}
\label{eq:7}
\end{equation}
As illustrated in \cref{fig:4}(b), when the object rotates around the center point, \(Z_{err}\) will fluctuate in tandem with \(h_{bbox} \), leading to an amplification effect on the distribution of \(H_{err}\). Similar to the squaring operation in the chi-square distribution, the multiplication in \cref{eq:7} introduces nonlinear transformation, resulting in a skewed data distribution. (2) Bounding box height error \(h_{err} \) : Compared with other geometry errors, \(h_{err} \) is located in the denominator of the modified \cref{eq:6}. Its gradient calculation formula can be expressed as:
\begin{equation}
\begin{aligned}
\frac{{\partial {L_{depth}}}}{{\partial {h_{err}}}} =  \frac{{\partial {L_{depth}}}}{{\partial Z}} \cdot \frac{{f \cdot H}}{{{{({h_{bbox}} - {h_{err}})}^2}}}
\end{aligned}
\label{eq:8}
\end{equation}
During the backpropagation process, the gradient calculations on \(h_{err} \) are non-linear and unstable, thus increasing the difficulty of model training and convergence.

In summary, we obtain \(Z_{geo}\) using \(h_{bbox}\) instead of \(h_c\) and select \(Z_{err}\) to compensate the projection formula. These measures can greatly optimize data distribution and reduce the model's learning costs. The difficulty of predicting depth-related attributes is ranked as follows:
\begin{equation}
Z > {h_{err}} > {H_{err}} > {Z_{err}} \approx H > > {h_{bbox}}
\end{equation}

\subsection{Loss Function}

The training loss consists of four distinct components: region loss \(L_{region}\), depth map loss \(L_{dmap}\), 2D loss \(L_{2D}\), and 3D loss \(L_{3D}\). Specifically, \(L_{2D}\) contains the losses of object category, bounding box, GIoU, and projected center point, while \(L_{3D}\) includes the losses of 3D size, orientation, and central depth. We denote the weights of 2D and 3D losses from \({\lambda _1}\) to \({\lambda _7}\). The overall loss is formulated as:
\begin{equation}
{L_{overall}} = {L_{2D}} + {L_{3D}} + {\lambda _8}{L_{dmap}} + {\lambda _9}\sum\limits_{i = 0}^4 {L_{region}^i} 
\end{equation}

\section{Experiments}


\subsection{Settings}
\boldparagraph{Dataset.}
We select the widely used KITTI benchmark~\cite{kitti} to evaluate our proposed model. This 3D object detection benchmark consists of 7481 training images and 7518 testing images. It has three difficulty levels (Easy, Moderate, and Hard) and three classes (Car, Pedestrian, and Cyclist). According to the prior work~\cite{chen20153d}, we divide the 7481 training images into a training set with 3712 images and a validation set with 3769 images for ablation study.

\boldparagraph{Evaluation Metrics.}
We present the detection outcomes categorized by three difficulty levels: easy, moderate, and hard. Consistent with prior works, our evaluation relies on average precision (AP) metrics for both 3D bounding boxes and bird's-eye view, referred to as \( AP_{3D} \) and \( AP_{BEV} \), which are calculated at 40 recall positions following the established protocol \cite{simonelli2019disentangling}. The benchmark ranks all methods based on the moderate \( AP_{3D} \) metrics of ``Car'' category.

\boldparagraph{Implementation Details.}
We take ResNet50~\cite{resnet} as our backbone. In regard to Transformer, we adopt 8 heads in each attention module and set the number of queries 50 with 4 sampling points in the multi-scale deformable attention. The weights of losses are set as \{2, 5, 2, 10, 1, 1, 1, 1, 1\} for \({\lambda _{1}}\) to \({\lambda _{9}}\). On a single RTX 3090 GPU, using Mixup3D \cite{monolss}, we train MonoDGP for 250 epochs with a batch size of 8 and a learning rate of 2e-4. We utilize AdamW~\cite{adamW} optimizer with weight decay and decrease learning rate by a factor of 0.5 at epochs 85, 125, 165, and 225. To accelerate the model's convergence, we adopt group-wise one-to-many assignment~\cite{groupdetr} with 11 groups. During inference, we filter out queries with category confidence lower than 0.2.


\begin{table}[t]
\centering
\resizebox{0.48\textwidth}{!}{
\begin{tabular}{l|c|c|c|c}
\hline
\multirow{2}{*}{Methods} & \multirow{2}{*}{\begin{tabular}[c]{@{}c@{}}Val, Mod.\\ $AP_{3D}$\end{tabular}} & \multirow{2}{*}{\begin{tabular}[c]{@{}c@{}}Params\\ (M)$\downarrow$\end{tabular}} & \multirow{2}{*}{\begin{tabular}[c]{@{}c@{}}FLOPs\\ (G)$\downarrow$\end{tabular}} & \multirow{2}{*}{\begin{tabular}[c]{@{}c@{}}Runtime\\ (ms)$\downarrow$\end{tabular}} \\
 &  &  &  &  \\ \hline
Baseline & 20.22 & \textbf{35.93} & \textbf{59.72} & \textbf{35} \\
MonoDGP (Ours) & \textbf{22.34} & 38.90 & 68.99 & 42 \\
w/o Decoupled Query & 21.48 & 37.54 & 65.75 & 39 \\
w/o RSH & 21.93 & 36.23 & 62.25 & 37 \\ \hline
\end{tabular} 
}
\caption{Ablation study of computational cost. We test the Runtime (ms) on a single RTX 3090 GPU with a batch size of 1.}
\label{tab:computation}
\end{table}

\subsection{Main Results}

To present the performance of our proposed method, we provide quantitative results on the KITTI test and val splits. 

As shown in \cref{tab:1}, we have compared advanced monocular 3D detectors in recent years. Without additional data, MonoDGP comprehensively surpasses the baseline model MonoDETR~\cite{monodetr} and achieves competitive results across most metrics. In particular, our method improves moderate \(AP_{3D}\) by up to 1.55\% on the test set and 1.73\% on the validation set, respectively. Except that easy \( AP_{BEV} \) on test set achieves the second-best lower than OccupancyM3D~\cite{occupancym3d}, which introduces extra data during the training phase, other metrics all reach the state-of-the-art performance.

It should be noted that we employ early stopping to select the best validation checkpoint, while we directly select the last few checkpoints for test submission. To mitigate the instability of the DETR-based method, all ablation studies follow the same protocol where models are trained 5 times on val set with median moderate \( AP_{3D} \) reported.

\subsection{Ablation Study}
\label{sec:abla}

\begin{table}

\centering
\resizebox{0.7\columnwidth}{!}{%
\begin{tabular}{c|ccc}
\hline
\multirow{2}{*}{\begin{tabular}[c]{@{}c@{}}Segment Embeddings \\ Threshold\end{tabular}} & \multicolumn{3}{c}{Val, $AP_{3D|R40}$} \\ \cline{2-4} 
 & Easy & Mod. & Hard \\ \hline
0.3 & 29.77 & 21.51 & 18.35 \\
0.5 & \textbf{30.57} & \textbf{21.64} & \textbf{18.74} \\
0.7 & 30.35 & 21.54 & 18.55 \\
w/o & 29.63 & 21.43 & 18.21 \\ \hline
\end{tabular}
}
\caption{The design of region segmentation head. ``w/o’' denotes enhancing visual and depth features without segment embeddings.}
\label{tab:2}
\end{table}

\boldparagraph{Computational cost.}
We analyze the computational costs of the main modules in \cref{tab:computation} (Depth Error does not incur additional computation). Although the decoupled query and RSH modules introduce a slight increase in computational overhead, they greatly improve the performance. Our complete network achieves a runtime of 42 ms, which is comparable to recent state-of-the-art methods such as MonoCD \cite{monocd} (36 ms) and OccupancyM3D \cite{occupancym3d} (112 ms).

\boldparagraph{Region segmentation head.}
For the multi-scale target region probabilities predicted by RSH, we set distinct thresholds to produce segment embeddings, which serve as input tokens for the depth encoder. The results are presented in \cref{tab:2}. The segment embeddings strength encoders' ability to comprehend the contextual relationships among image pixels. Experiments further highlight the critical role of segment embeddings. Moreover, the threshold settings also influence 3D evaluation metrics. A threshold of 0.5 achieves superior results compared to other thresholds.

\begin{table}

\centering
\resizebox{0.75\columnwidth}{!}{%
\begin{tabular}{l|ccc}
\hline
\multirow{2}{*}{Depth Predition Mode} & \multicolumn{3}{c}{Val, $AP_{3D|R40}$} \\ \cline{2-4} 
 & Easy & Mod. & Hard \\ \hline
Depth Map & 24.33 & 18.87 & 15.31 \\
Direct Depth & 26.60 & 19.23 & 16.47 \\ 
Geometric Depth ($h_{bbox}$) & 9.15 & 7.99 & 5.52 \\
Geometric Depth ($h_{c}$) & 27.63 & 20.73 & 18.04 \\
Weighted Fusion & 30.57 & 21.64 & 18.74 \\ \hline
GD + Bbox Height Error & 28.45 & 21.18 & 17.91 \\
GD + Dim Height Error & 30.46 & 22.09 & 18.85 \\
GD + Depth Error & \textbf{30.76} & \textbf{22.34} & \textbf{19.01} \\ \hline

\end{tabular}}
\caption{Ablation study of different depth prediction modes on KITTI val set. The weighted fusion mode integrates the results from the depth map, direct depth, and geometric depth ($h_{bbox}$). ``GD'' denotes the geometric depth calculated by the formula \( Z_{geo} = f \times H_{3D} / h_{bbox} \).}
\label{tab:3}
\end{table}

\boldparagraph{Depth prediction mode.}
Based on the same network, we compare different depth prediction modes in \cref{tab:3}. The experiment indicates that single depth predictions, whether using direct depth, depth map, or geometric depth ($h_{bbox}$ or $h_{c}$), generally perform poorly, especially for geometric depth ($h_{bbox}$). Considering only perspective transformation on the visual surface, the moderate \( AP_{3D}\) for geometric depth ($h_{bbox}$) is just 7.99\% (all \( AP_{3D}\) mentioned below default to the moderate level), significantly lower than other prediction modes. By fusing direct depth, depth map and geometric depth, the \( AP_{3D}\) improves to 21.64\%. When geometry errors are added to compensate for the projection formula, it can be observed that the performance achieves a substantial increase. The depth error wins the best performance with an \( AP_{3D}\) of 22.34\%. The dimension height error comes second at 22.09\% due to its skewed distribution. The bounding box height error only acquires 21.18\% in \( AP_{3D}\), obviously inferior to other geometry errors. This discrepancy can be attributed to the non-linearity of the gradient backpropagation calculation.

\begin{table}

\centering
\resizebox{\columnwidth}{!}{%
\begin{tabular}{cccc|ccc}
\hline
\multirow{2}{*}{\small Mixup3D} & 
\multirow{2}{*}{\begin{tabular}[c]{@{}c@{}} \small Decoupled\\ \small Query\end{tabular}} &
\multirow{2}{*}{\begin{tabular}[c]{@{}c@{}} \small Region\\ \small Segment\end{tabular}} & 
\multirow{2}{*}{\begin{tabular}[c]{@{}c@{}} \small Geometric\\ \small Depth Error\end{tabular}} & \multicolumn{3}{c}{Val, $AP_{3D|R40}$} \\ \cline{5-7} 
 &  &  &  & Easy & Mod. & Hard \\ \hline
\usym{2715} & \usym{2715} & \usym{2715} & \usym{2715} & 26.56 & 20.22 & 17.21 \\
\Checkmark & \usym{2715} & \usym{2715} & \usym{2715} & 28.17 & 20.79 & 17.52 \\

\Checkmark & \Checkmark & \usym{2715} & \usym{2715} & 29.82 & 21.38 & 18.04 \\
\Checkmark & \Checkmark & \Checkmark & \usym{2715} & 30.57 & 21.64 & 18.74 \\
\Checkmark & \Checkmark & \usym{2715} & \Checkmark & 30.08 & 21.93 & 18.87 \\
\Checkmark & \usym{2715} & \Checkmark & \Checkmark & 28.91 & 21.48 & 18.43 \\

\Checkmark & \Checkmark & \Checkmark & \Checkmark & \textbf{30.76} & \textbf{22.34} & \textbf{19.01} \\ \hline
\multicolumn{4}{c|}{Improvement} & + 4.20 & +2.12 & +1.80 \\ \hline
\end{tabular} }
\caption{The main ablation study.}
\label{tab:4}
\end{table}

\boldparagraph{Overall performance.}
We explore the impact of each improvement technique on the baseline. The main ablation study is shown in \cref{tab:4}. The MonoDGP employs four key techniques: mixup3D augmentation, decoupling and initialization of 2D object queries, region enhancement with segment embeddings, and depth error prediction based on geometric depth. Compared to the baseline, the \( AP_{3D}\) under three-level difficulties gain an increase by +4.20\%, +2.12\% and +1.80\%, respectively. These experiments validate the effectiveness of each technique utilized in MonoDGP.
\section{Conclusion}
In this paper, we propose a novel transformer-based monocular method called MonoDGP, which adopts geometry errors to correct the projection formula. We also comprehensively analyze the properties and effectiveness of geometry errors, from the perspectives of marginal feature extraction, data distribution, and gradient propagation. Error prediction is expected to become a better alternative to multi-depth prediction, without complex multi-branch implementations. Furthermore, MonoDGP introduces a 2D visual decoder for query initialization and a region segmentation head for feature enhancement. These improvements have achieved superior performance on the KITTI benchmark. We can also extend our approach to dense depth map prediction for the target region, by only considering errors between geometric depth and surface points of the objects.

\clearpage
{
   \small
   \bibliographystyle{ieeenat_fullname}
   \bibliography{main}
}

\clearpage

\appendix
\maketitlesupplementary

\section{Detailed Discussion on Depth Error}
\label{sec:depth_error}
In our method, we regard the distance between the camera plane and the car's closest wheel point as the geometric depth. However, this assumption is appropriate when the camera has the same height of the object. The height inconsistency will lead to the bias $l_{bias}$ between the actual geometric depth $Z_{geo}$ and the wheel depth $Z_w$. We set the height ratio $\gamma$ of the camera height $H_{cam}$ to the object height $H$ as follows:
\begin{equation}
\gamma = \frac{H_{cam}}{H}
\label{eq:a1}
\end{equation}

We will discuss how the height ratio affects the distribution of the depth error $Z_{err}$.  

\begin{figure}[htbp]
   \centering
   \includegraphics[width=1.0\linewidth]{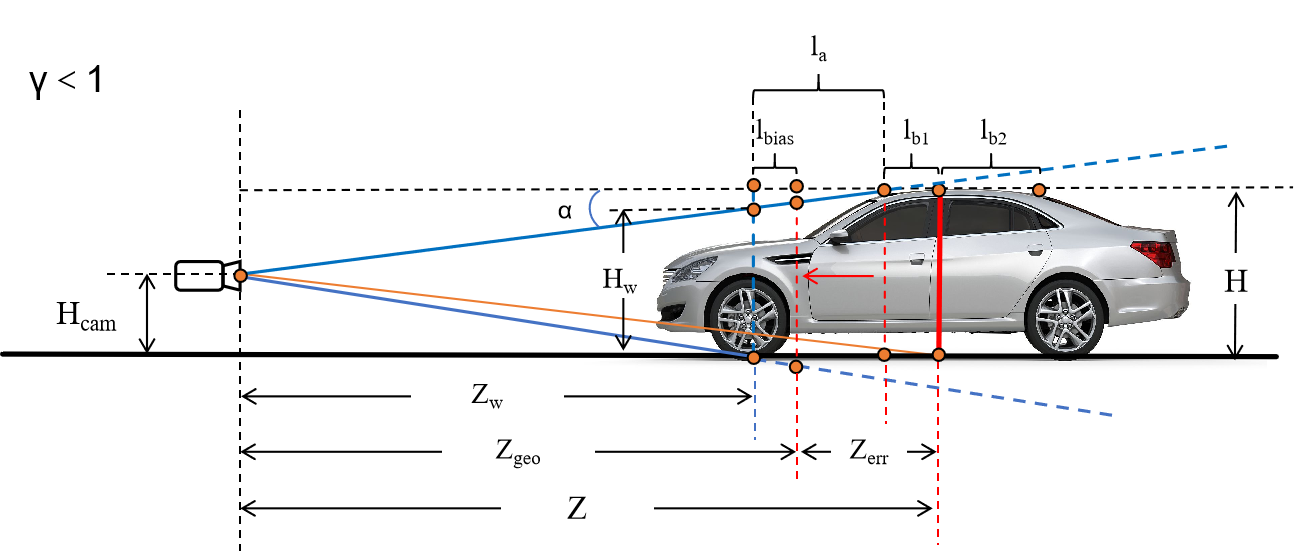}
   \caption{The perspective transformation when the camera height is lower than the object height.}
   \label{fig:a}
\end{figure}

The vehicle is treated as a trapezoid. The closest wheel locates in the lowest position of the 2D bounding box, while the highest position in the object surface will change with the height ratio. As shown in \cref{fig:a}, when $\gamma < 1$, the wheel depth is shorter than the geometric depth, which can be expressed as:
\begin{equation}
Z_{geo} = Z_{w} + l_{bias}
\label{eq:a2}
\end{equation}

To calculate the wheel bias, we first represent the height at the wheel point based on the similar triangle theory:
\begin{equation}
tan\alpha = \frac{H-H_{cam}}{Z_w+l_a}=\frac{H-H_w}{l_a}
\label{eq:a3}
\end{equation}

\begin{equation}
H_w = H - \frac{(H-H_{cam})\cdot l_a}{Z_w + l_a}
\label{eq:a4}
\end{equation}

And then we utilize $H_w$ to compute $l_{bias}$ :
\begin{equation}
\frac{H}{Z_{geo}}=\frac{H_w}{Z_w}
\label{eq:a5}
\end{equation}

\begin{equation}
l_{bias} = \frac{(H-H_{cam}) \cdot Z_w \cdot l_a }{H \cdot Z_w + H_{cam} \cdot l_a} =  \frac{(1-\gamma) \cdot Z_w \cdot l_a }{Z_w + \gamma \cdot l_a}
\label{eq:a6}
\end{equation}

We can express depth error as follows:

\begin{equation}
Z_{err} = l_{b1} + l_a - l_{bias} = l_{b1} + \sigma_1 \cdot l_a
\label{eq:a7}
\end{equation}

\begin{equation}
\sigma_1 = \frac{\gamma }{1-\frac{(1-\gamma )\cdot l_a}{Z_w + l_a}}
\label{eq:a8}
\end{equation}
where $\gamma<\sigma_1 < 1$, $l_{bias} < (1-\gamma) \cdot l_a$. The original depth error, which should be perspective-invariant, is calculated by the formula $Z_{err} = l_{b1} + l_a$. Except for the vehicle's own attributes, $\gamma$ and $Z_w$ also affect the depth error. The closer $\sigma_1$ is to 1, the less effect it has. According to the \cref{eq:a8}, $\sigma_1$ reduces as $Z_w$ increases and $\gamma$ decreases.

\begin{figure}[htbp]
   \centering
   \includegraphics[width=1.0\linewidth]{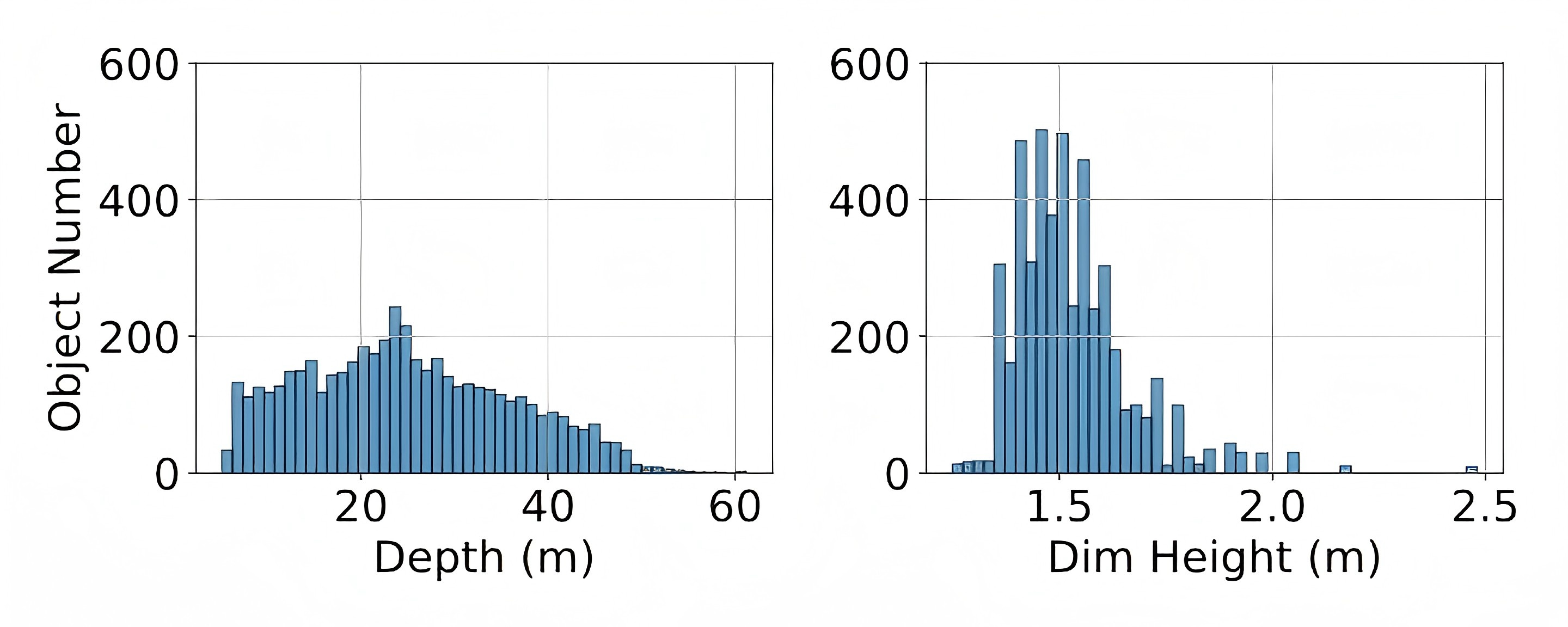}
   \caption{The data distribution of the object's central depth and dimension height on the KITTI training set.}
   \label{fig:b}
\end{figure}

To present the greatest impact of the height ratio, we take an extreme example based on the \cref{fig:b}, and make $\sigma_1$ as smaller as possible. Specifically, we set $H_{cam}=1.5 m$, $H=1.8 m$, $\gamma=\frac{5}{6}$, $l_a=1 m$, $Z_w=50 m$. From the \cref{eq:a6} and \cref{eq:a8}, we obtain $\sigma_1 \approx 0.84$ and $l_{bias} \approx 0.16m$. This extreme bias value is significantly lower than the whole depth value.

\begin{figure}[htbp]
   \centering
   \includegraphics[width=1.0\linewidth]{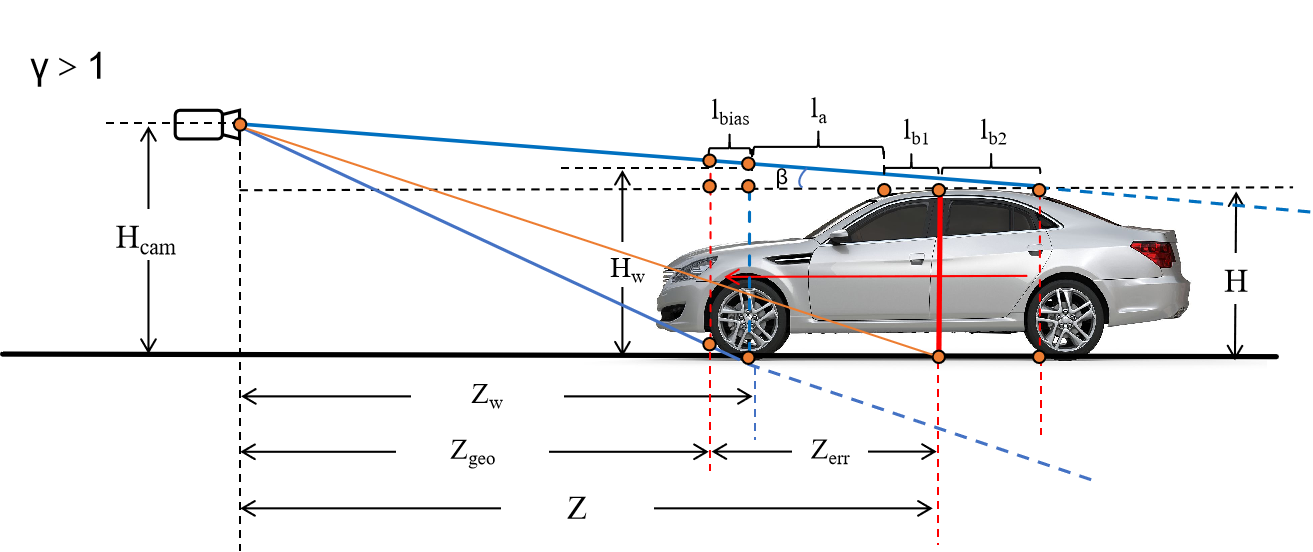}
   \caption{The perspective transformation when the camera height is higher than the object height.}
   \label{fig:c}
\end{figure}

As shown in \cref{fig:c}, when $\gamma >1$, the wheel depth is larger than the geometric depth, which can be expressed as:

\begin{equation}
Z_{geo} = Z_w - l_{bias}
\label{eq:a9}
\end{equation}

We can achieve the wheel bias similar to the previous proof process:

\begin{equation}
tan\beta = \frac{H_{cam}-H}{Z_w + l_a + l_{b1} + l_{b2}} = \frac{H_w - H}{l_a + l_{b1} + l_{b2}}
\label{eq:a10}
\end{equation}

\begin{equation}
H_w = \frac{H\cdot Z_w +H_{cam} \cdot (l_a + l_{b1} + l_{b2} )}{Z_w + l_a + l_{b1} +l_{b2}}
\label{eq:a11}
\end{equation}

\begin{equation}
l_{bias} = \frac{(\gamma - 1) \cdot Z_w \cdot (l_a + l_{b1} + l_{b2})}{Z_w + \gamma \cdot (l_a + l_{b1} + l_{b2})}
\label{eq:a12}
\end{equation}

Homogeneously, we can express depth error as follows:

\begin{equation}
Z_{err}  = l_{b1} + l_a + \sigma_2 \cdot (l_a + l_{b1} + l_{b2} )
\label{eq:a13}
\end{equation}

\begin{equation}
\sigma_2 = \frac{(\gamma -1)}{1+\frac{\gamma  \cdot (l_a + l_{b1} + l_{b2} )}{Z_w}}
\label{eq:a14}
\end{equation}
where $0< \sigma_2 < \gamma -1$, $l_{bias} <(\gamma -1) \cdot (l_a+l_{b1}+l_{b2})$. The closer $\sigma_2$ is to 0, the less effect it has. According to the \cref{eq:a14}, $\sigma_2$ increases as $Z_w$ and $\gamma$ increase. 

To show the height ratio's maximum impact, we also suppose an extreme case and make $\sigma_2$ as larger as possible. To be more specific, we set $H_{cam}=1.5 m$, $H=1.25 m$, $\gamma=\frac{6}{5}$, $l_a+l_{b1}+l_{b2}=2 m$, $Z_w=50 m$. Based on the \cref{eq:a12} and \cref{eq:a14}, we achieve $\sigma_2 \approx 0.19$ and $l_{bias} \approx 0.38m$. This bias value is higher than the value calculated when $\gamma < 1$, but has a slight effect on the whole depth.

In most instances, the camera height is close to the vehicle height, which means $\gamma \approx 1$ and the depth error is roughly perspective-invariant for the car category. Even if the object height is obviously different from the camera height, the network can directly learn and predict this tiny bias compared with the whole depth. The error prediction is still a simple and effective method to replace the multi-depth prediction.

\section{Discussion on Geometric Constraints}

Previous works like Deep3DBox~\cite{deep3dbox} and Shift R-CNN~\cite{shift-rcnn} enforce strict geometric constraints by tightly fitting projections of the 3D bounding box into the 2D box. While recent methods such as MonoGR2~\cite{monogr2} and GUPNet~\cite{gupnet} formulate constraints based on geometric similarity, where under vehicle-mounted camera perspectives and fixed focal length, the object's center depth can be uniquely determined through the proportional relationship between its 3D height and 2D projected height. 

Projection-alignment constraints exhibit quadratic errors from 2D boundary localization inaccuracies, while height-ratio constraints demonstrate linear errors confined to height predictions. The former fails with truncated objects requiring full 2D contours, whereas the latter maintains functionality under partial occlusions through visible height segments. Height-ratio constraints surpass projection-alignment methods in stability (linear vs. quadratic errors), efficiency (closed-form vs. iterative), and robustness (partial vs. full contours), establishing them as core geometric priors for monocular 3D detection. Future frameworks could incorporate projection-alignment constraints as auxiliary regularizers within joint optimization.

\section{Detailed Loss Function}

The 2D loss $L_{2D}$ adopts focal loss~\cite{lin2017focal} to estimate the object categories, L1 loss to estimate the projected center $(x_{3d},y_{3d})$ and 2D sizes $(l,r,t,b)$, and GIoU loss for the bounding box. We can formulate the 2D object loss as:
\begin{equation}
L_{2D} = \lambda_1 L_{cls} +\lambda_2 L_{2dsize} +\lambda_3 L_{xy} +\lambda_4 L_{giou}
\label{eq:a15}
\end{equation}

The 3D loss follows MonoDLE~\cite{monodle} to predict 3D sizes $(h_{3d},w_{3d},l_{3d})$ and orientation angle $\alpha$. As for the depth prediction, an uncertainty regression loss based on the Laplacian distribution is defined as:

\begin{equation}
{L_{depth}} = \frac{{\sqrt 2 }}{{{\sigma _d}}}{\left\| { \frac{{f \cdot H}}{{{h_{bbox}}}} + {Z_{err}} - {Z_{gt}}} \right\|_1} + \log ({\sigma _d})
\label{eq:a16}
\end{equation}
where $\sigma_d$ is the standard deviation of the distribution.

We can formulate 3D object loss as:
\begin{equation}
L_{3D} = \lambda_5 L_{3dsize} +\lambda_6 L_{angle} +\lambda_7 L_{depth}
\label{eq:a17}
\end{equation}

The depth map loss $L_{dmap}$ utilizes focal loss to predict categorical foreground depth map. More detailed information about $L_{dmap}$ can be found in MonoDETR~\cite{monodetr}.

\section{Experiments on Other Categories}
Since segment embeddings are mainly trained to distinguish between background and target, they can easily handle multiple classes without modification. Ablation studies of other categories are shown in \cref{tab:a}. In particular, error prediction significantly improves 3D pedestrian prediction compared to cars and cyclists. This can be explained that pedestrians have consistent depth errors across orientations, whereas cyclists have irregular shapes and greater geometric uncertainty. These spatial uncertainties may degrade the effectiveness of the projection transformation.

We also compare the pedestrian and cyclist detection results in \cref{tab:b}. Specifically, our method achieves a superior performance on all levels of difficulty for pedestrian detection, benefiting from its simple and stable geometric structures. However, the performance for the cyclist category falls short of the best. 

Notably, despite these geometric challenges, our cyclist detection performance remains competitive among methods without extra training data. This underscores the generalizability of RSH module for complex categories.

\begin{table}[t]
\centering

\resizebox{1.0\columnwidth}{!}{
\begin{tabular}{l|cccccc}
\hline
\multirow{3}{*}{Methods} & \multicolumn{6}{c}{Val, IoU=0.5, $AP_{3D|R40}$} \\ \cline{2-7} 
 & \multicolumn{3}{c|}{Pedestrian} & \multicolumn{3}{c}{Cyclist} \\ \cline{2-7} 
 & Easy & Mod. & \multicolumn{1}{c|}{Hard} & Easy & Mod. & Hard \\ \hline
MonoDGP (Ours) & \textbf{13.77} & \textbf{10.06} & \multicolumn{1}{c|}{\textbf{7.96}} & \textbf{12.21} & \textbf{6.61} & \textbf{5.95} \\
w/o Segment Embeddings & 13.02 & 9.67 & \multicolumn{1}{c|}{7.66} & 10.56 & 5.22 & 4.68 \\
w/o RSH & 12.50 & 9.42 & \multicolumn{1}{c|}{7.34} & 9.16 & 4.34 & 4.18 \\
w/o Depth Error & 9.90 & 7.55 & \multicolumn{1}{c|}{6.09} & 11.13 & 5.86 & 5.51 \\ \hline
\end{tabular}
}
\caption{Ablation study of the pedestrian and cyclist categories on the KITTI val set.}
\label{tab:a}
\end{table}

\begin{table}[t]
\centering
\resizebox{1.0\columnwidth}{!}{%
\begin{tabular}{l|c|cccccc}
\hline
\multirow{3}{*}{Methods} & \multirow{3}{*}{\begin{tabular}[c]{@{}c@{}}Extra \\ data\end{tabular}} & \multicolumn{6}{c}{Test, IoU=0.5, $AP_{3D|R40}$ } \\ \cline{3-8} 
 &  & \multicolumn{3}{c|}{Pedestrian} & \multicolumn{3}{c}{Cyclist} \\ \cline{3-8} 
 &  & Easy & Mod. & \multicolumn{1}{c|}{Hard} & Easy & Mod. & Hard \\ \hline
CaDDN~\cite{caddn} & \multirow{2}{*}{LiDAR} & 12.87 & 8.14 & \multicolumn{1}{c|}{6.76} & {\ul 7.00} & 3.41 & {\ul 3.30} \\
OccupancyM3D~\cite{occupancym3d} &  & 14.68 & 9.15 & \multicolumn{1}{c|}{7.80} & \textbf{7.37} & {\ul 3.56} & 2.84 \\ \hline
MonoPGC~\cite{monopgc} & Depth & 14.16 & {\ul 9.67} & \multicolumn{1}{c|}{{\ul 8.26}} & 5.88 & 3.30 & {\ul 2.85} \\ \hline
GUPNet~\cite{gupnet} & \multirow{4}{*}{None} & {\ul 14.72} & 9.53 & \multicolumn{1}{c|}{7.87} & 4.18 & 2.65 & 2.09 \\
MonoCon~\cite{monocon} &  & 13.10 & 8.41 & \multicolumn{1}{c|}{6.94} & 2.80 & 1.92 & 1.55 \\
DEVIANT~\cite{deviant} &  & 13.43 & 8.65 & \multicolumn{1}{c|}{7.69} & 5.05 & 3.13 & 2.59 \\
MonoDDE~\cite{monodde} &  & 11.13 & 7.32 & \multicolumn{1}{c|}{6.67} & 5.94 & \textbf{3.78} & \textbf{3.33} \\
MonoDETR~\cite{monodetr} &  & 12.65 & 7.19 & \multicolumn{1}{c|}{6.72} & 5.12 & 2.74 & 2.02 \\ \hline
MonoDGP (Ours) & None & \textbf{15.04} & \textbf{9.89} & \multicolumn{1}{c|}{\textbf{8.38}} & 5.28 & 2.82 & 2.65 \\ \hline
\end{tabular}
}
\caption{Comparisons of the pedestrian and cyclist categories on the KITTI test set. We \textbf{bold} the best results and \underline{underline} the second-best results.}
\label{tab:b}
\end{table}




\section{Sensitivity to Initial Features}
Since error prediction mode heavily relies on good geometric features, the inaccuracies of initial features can significantly impact the convergence and performance of the proposed network. The Initial features, such as 3D dimension height ($H_{3D}$) and 2D bounding box height ($h_{bbox}$), are crucial for geometric depth calculation. To analyze their individual impacts, we conduct sensitivity experiments replacing predicted $H_{3D}$ and $h_{bbox}$ with ground truth values. 

As shown in \cref{tab:d}, the perfectly accurate geometric depth improves moderate $AP_{3D}$ by up to 26.21\%, highlighting the significance of these features. Compared to $h_{bbox}$, the network is more sensitive to $H_{3D}$ errors due to its inherent difficulty as a 3D property. There also exists a coupling relationship between $h_{bbox}$ and $H_{3D}$. Simultaneously replacing both features with ground truth values performs much better than replacing them individually. Current limitations mainly arise from height prediction error accumulation in the perspective projection. Improvements in monocular features, particularly for $H_{3D}$, will further enhance the performance of error prediction in the future.

\begin{table}[t]
\centering

\resizebox{0.48\textwidth}{!}{
\begin{tabular}{cc|ccc}
\hline
\multicolumn{2}{c|}{Geometric Depth ( \( f \times H_{3D} / h_{bbox} \) )} & \multicolumn{3}{c}{Val, IoU=0.7, $AP_{3D|R40}$} \\ \hline
\multicolumn{1}{c|}{Ground Truth \(H_{3D}\)} & Ground Truth \(h_{bbox}\) & Easy & Mod. & Hard \\ \hline
\multicolumn{1}{c|}{\usym{2715}} & \usym{2715} & 30.76 & 22.34 & 19.01 \\
\multicolumn{1}{c|}{\Checkmark} & \usym{2715} & 39.10 & 31.89 & 27.51 \\
\multicolumn{1}{c|}{\usym{2715}} & \Checkmark & 33.62 & 25.04 & 21.97 \\
\multicolumn{1}{c|}{\Checkmark} & \Checkmark & \textbf{57.81} & \textbf{48.55} & \textbf{41.92} \\ \hline
\end{tabular}
}
\caption{Sensitivity study on the KITTI val set for the car category.}
\label{tab:d}
\end{table}

\begin{table}[t]
\centering
\resizebox{0.48\textwidth}{!}{
\begin{tabular}{l|ccc}
\hline
\multicolumn{1}{l|}{\multirow{2}{*}{Depth Prediction Mode}} & \multicolumn{3}{c}{Val, IoU=0.7, $AP_{3D|R40}$} \\ \cline{2-4} 
\multicolumn{1}{c|}{} & Easy & Mod. & Hard \\ \hline
Direct Depth & \multicolumn{1}{l}{24.33} & \multicolumn{1}{l}{18.87} & \multicolumn{1}{l}{15.31} \\
DAv2-small (HS) + Depth Error & 11.45 & 9.11 & 7.86 \\
DAv2-small (VK2) + Depth Error & 27.04 & 20.48 & 17.52 \\
DAv2-base (VK2) + Depth Error & \textbf{27.86} & \textbf{21.15} & \textbf{18.10} \\ \hline
\end{tabular}
}
\caption{Ablation Study of pre-trained MDE. `DAv2' denotes Depth Anything V2 \cite{depthanythingv2} method, `HS' denotes pre-trained on indoor dataset Hypersim \cite{hypersim}, `VK2' denotes pre-trianed on outdoor dataset Virtual KITTI 2 \cite{VK2}.}
\label{tab:e}
\end{table}

\begin{table*}[!ht]
\centering
\resizebox{0.90\textwidth}{!}{%
\begin{tabular}{c|l|c|cccc|cccc}
\hline
\multirow{2}{*}{Difficulty} & \multirow{2}{*}{Methods} & \multirow{2}{*}{Extra} & \multicolumn{4}{c|}{$AP_{3D}$} & \multicolumn{4}{c}{$APH_{3D}$} \\ 
\cline{4-11} 
 &  &  & All & 0-30 & 30-50 & 50-$\infty$ & All & 0-30 & 30-50 & 50-$\infty$ \\ 
\hline
\multirow{8}{*}{{Level\_1 \newline (IoU=0.7)}} 
 & CaDDN \cite{caddn} & LiDAR & 5.03 & 15.54 & 1.47 & 0.10 & 4.99 & 14.43 & 1.45 & 0.10 \\
 & PatchNet \cite{patchnet} in \cite{pct} & Depth & 0.39 & 1.67 & 0.13 & 0.03 & 0.39 & 1.63 & 0.12 & 0.03 \\
 & PCT \cite{pct} & Depth & 0.89 & 3.18 & 0.27 & 0.07 & 0.88 & 3.15 & 0.27 & 0.07 \\
 & M3D-RPN \cite{m3d-rpn} in \cite{caddn} & None & 0.35 & 1.12 & 0.18 & 0.02 & 0.34 & 1.10 & 0.18 & 0.02 \\
 & GUPNet \cite{gupnet} in \cite{deviant} & None & 2.28 & 6.15 & 0.81 & 0.03 & 2.27 & 6.11 & 0.80 & 0.03 \\
 & DEVIANT \cite{deviant} & None & 2.69 & 6.95 & {\ul 0.99} & 0.02 & 2.67 & 6.90 & {\ul 0.98} & 0.02 \\
 & MonoUNI \cite{monouni} & None & {\ul 3.20} & {\ul 8.61} & 0.87 & {\ul 0.13} & {\ul 3.16} & {\ul 8.50} & 0.86 & {\ul 0.12} \\
 & \textbf{MonoDGP (Ours)} & None & \textbf{4.28} & \textbf{10.24} & \textbf{1.15} & \textbf{0.16} & \textbf{4.23} & \textbf{10.10} & \textbf{1.14} & \textbf{0.16} \\ 
\hline
\multirow{8}{*}{{Level\_2\newline (IoU=0.7)}} 
 & CaDDN \cite{caddn} & LiDAR & 4.49 & 14.50 & 1.42 & 0.09 & 4.45 & 14.38 & 1.41 & 0.09 \\
 & PatchNet \cite{patchnet} in \cite{pct} & Depth & 0.38 & 1.67 & 0.13 & 0.03 & 0.36 & 1.63 & 0.11 & 0.03 \\
 & PCT \cite{pct} & Depth & 0.66 & 3.18 & 0.27 & 0.07 & 0.66 & 3.15 & 0.26 & 0.07 \\
 & M3D-RPN \cite{m3d-rpn} in \cite{caddn} & None & 0.35 & 1.12 & 0.18 & 0.02 & 0.33 & 1.10 & 0.17 & 0.02 \\
 & GUPNet \cite{gupnet} in \cite{deviant} & None & 2.14 & 6.13 & 0.78 & 0.02 & 2.12 & 6.08 & 0.77 & 0.02 \\
 & DEVIANT \cite{deviant} & None & 2.52 & 6.93 & {\ul 0.95} & 0.02 & 2.50 & 6.87 & {\ul 0.94} & 0.02 \\
 & MonoUNI \cite{monouni} & None & {\ul 3.04} & {\ul 8.59} & 0.85 & {\ul 0.12} & {\ul 3.00} & {\ul 8.48} & 0.84 & {\ul 0.12} \\
 & \textbf{MonoDGP (Ours)} & None & \textbf{4.00} & \textbf{10.20} & \textbf{1.13} & \textbf{0.15} & \textbf{3.96} & \textbf{10.08} & \textbf{1.12} & \textbf{0.15} \\ 
\hline
\multirow{8}{*}{{Level\_1\newline (IoU=0.5)}} 
 & CaDDN \cite{caddn} & LiDAR & 17.54 & 45.00 & 9.24 & 0.64 & 17.31 & 44.46 & 9.11 & 0.62 \\
 & PatchNet \cite{patchnet} in \cite{pct} & Depth & 2.92 & 10.03 & 1.09 & 0.23 & 2.74 & 9.75 & 0.96 & 0.18 \\
 & PCT \cite{pct} & Depth & 4.20 & 14.70 & 1.78 & 0.39 & 4.15 & 14.54 & 1.75 & 0.39 \\
 & M3D-RPN \cite{m3d-rpn} in \cite{caddn} & None & 3.79 & 11.14 & 2.16 & 0.26 & 3.63 & 10.70 & 2.09 & 0.21 \\
 & GUPNet \cite{gupnet} in \cite{deviant} & None & 10.02 & 24.78 & 4.84 & 0.22 & 9.94 & 24.59 & 4.78 & 0.22 \\
 & DEVIANT \cite{deviant} & None & {\ul 10.98} & {\ul 26.85} & {\ul 5.13} & 0.18 & {\ul 10.89} & {\ul 26.64} & {\ul 5.08} & 0.18 \\
 & MonoUNI \cite{monouni} & None & {\ul 10.98} & 26.63 & 4.04 & {\ul 0.57} & 10.73 & 26.30 & 3.98 & {\ul 0.55} \\
 & \textbf{MonoDGP (Ours)} & None & \textbf{12.36} & \textbf{31.12} & \textbf{5.78} & \textbf{1.24} & \textbf{12.18} & \textbf{30.68} & \textbf{5.71} & \textbf{1.22} \\ 
\hline
\multirow{8}{*}{{Level\_2\newline (IoU=0.5)}} 
 & CaDDN \cite{caddn} & LiDAR & 16.51 & 44.87 & 8.99 & 0.58 & 16.28 & 44.33 & 8.86 & 0.55 \\
 & PatchNet \cite{patchnet} in \cite{pct} & Depth & 2.42 & 10.01 & 1.07 & 0.22 & 2.28 & 9.73 & 0.97 & 0.16 \\
 & PCT \cite{pct} & Depth & 4.03 & 14.67 & 1.74 & 0.36 & 4.15 & 14.51 & 1.71 & 0.35 \\
 & M3D-RPN \cite{m3d-rpn} in \cite{caddn} & None & 3.61 & 11.12 & 2.12 & 0.24 & 3.46 & 10.67 & 2.04 & 0.20 \\
 & GUPNet \cite{gupnet} in \cite{deviant} & None & 9.39 & 24.69 & 4.67 & 0.19 & 9.31 & 24.50 & 4.62 & 0.19 \\
 & DEVIANT \cite{deviant} & None & 10.29 & {\ul 26.75} & {\ul 4.95} & 0.16 & 10.20 & {\ul 26.54} & {\ul 4.90} & 0.16 \\
 & MonoUNI \cite{monouni} & None & {\ul 10.38} & 26.57 & 3.95 & {\ul 0.53} & {\ul 10.24} & 26.24 & 3.89 & {\ul 0.51} \\
 & \textbf{MonoDGP (Ours)} & None & \textbf{11.71} & \textbf{31.02} & \textbf{5.61} & \textbf{1.17} & \textbf{11.56} & \textbf{30.58} & \textbf{5.54} & \textbf{1.15} \\ 
\hline 
\end{tabular} 
}
\caption{Results on the Waymo val set for the vehicle category. Compared with methods without extra data, we \textbf{bold} the best results and {\ul underline} the second-best results.}
\label{tab:c}
\end{table*}



\section{Initial Depth from Pre-trained MDE}
Monocular depth estimation (MDE) models have developed for many years. We can also utilize the pre-trained MDE to provide a roughly approximate surface depth, similar to geometric depth, which may render the learning problem even simpler.

To explore this possibility, we exploit Depth Anything V2 \cite{depthanythingv2} to generate depth maps. Based on the initial metric depth, error prediction can achieve better performance compared to direct prediction in \cref{tab:a}. However, MDE heavily relies on pre-trained datasets, while geometric depth relies on its own attributes without additional parameters. This will limit the generalization of achieving initial depth from pre-trained MDE.


\section{Experiments on Waymo Open Dataset}
Waymo \cite{waymo} evaluates objects at Level\_1 and Level\_2, which are determined by the number of LiDAR points within their 3D bounding boxes. The experiments is conducted across three distance ranges: [0, 30), [30, 50), and [50, $\infty$) meters. Performance on the Waymo dataset is assessed by average precision $AP_{3D}$ and average precision weighted by heading $APH_{3D}$.

We follow the DEVIANT \cite{deviant} split to generate 52,386 training and 39,848 validation images by sampling every third frame.  For fairness, we mainly compare with methods using the same split in \cref{tab:c}. Our method achieves state-of-the-art performance without extra data across all ranges, particularly for distant objects. These results further validate the effectiveness and generalizability of MonoDGP. It is worth noting that CaDDN \cite{caddn}'s performance is better than MonoDGP, this discrepancy may be attributed to different dataset splits and introduction of LiDAR data.

\section{Qualitative Discussion and Visualization}
\begin{figure*}
   \centering
   \includegraphics[width=0.96\linewidth, height=21.3cm]{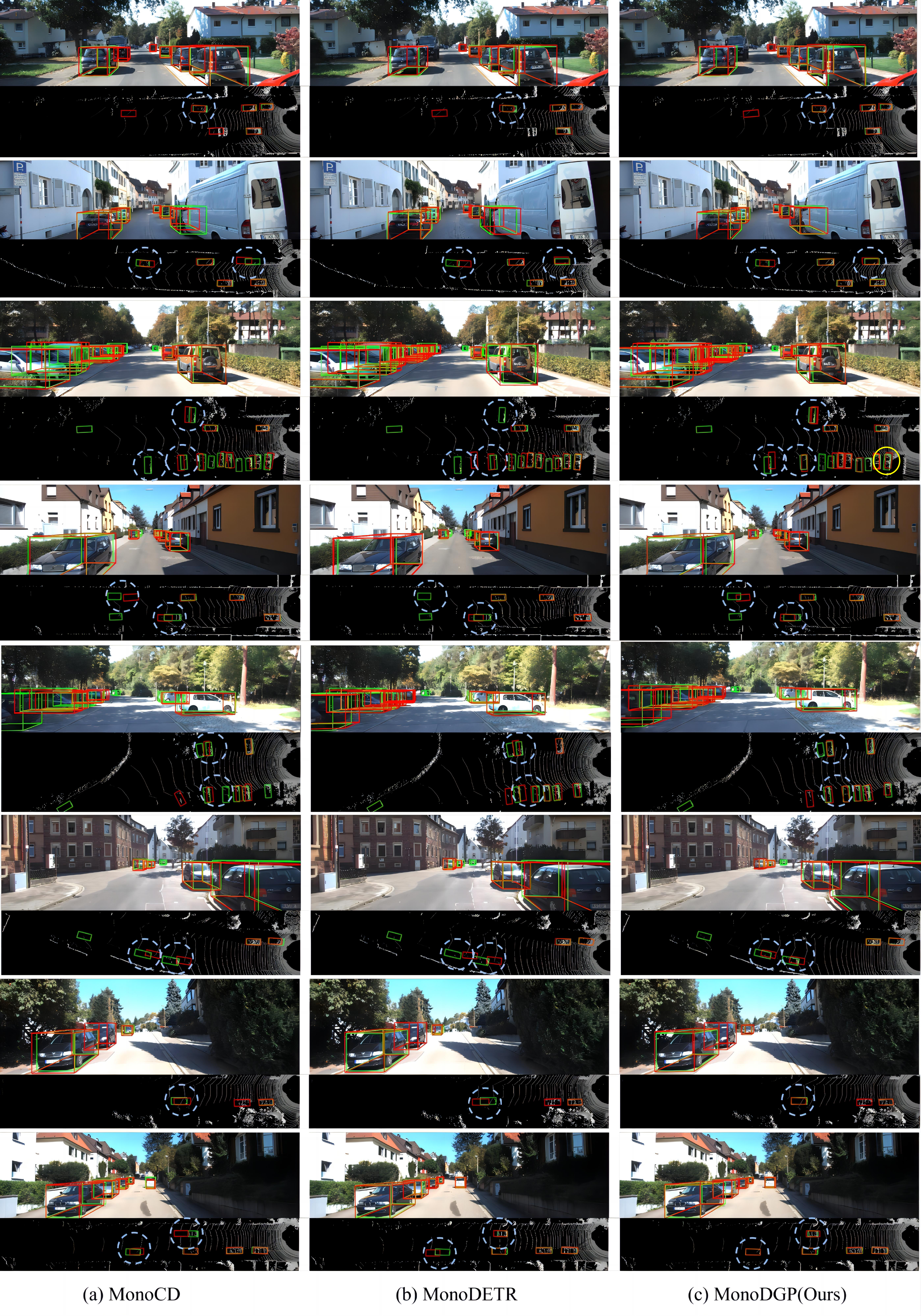}
   \caption{Qualitative results on KITTI validation set. (a) MonoCD (b) MonoDETR (c) MonoDGP (ours). In each group of images, the first row shows the camera view, and the second row shows the bird's-eye view. {\color{green} Green} represents the ground truth of boxes, while {\color{red} Red} represents the prediction results. We also circle some objects to highlight the difference between the baseline model and our method.}
   \label{fig:d}
\end{figure*}

To provide a more intuitive comparison between our method and the baseline models, we visualize some 3D detection results from both the camera view and the bird's-eye view on the KITTI validation set. As shown in \cref{fig:d}, our method demonstrates superior performance on distant and length-occluded objects.

However, since error prediction is affected by the initial accuracy of geometric depth, which is calculated from height relationships, height occlusion remains a challenge for our method. For the leftmost vehicle in the third example of \cref{fig:d}, bushes block out its lower part, weakening the accuracy of height and consequently propagating errors to depth prediction. This failure case highlights the need for further improvements in handling height occlusion, potentially through the integration of additional contextual information or more robust occlusion-aware models.

\end{document}